%% file: main.tex
\documentclass{article}
\usepackage[numbers]{natbib}
\usepackage{PRIMEarxiv}

\usepackage[hidelinks]{hyperref}
\usepackage{multirow}
\usepackage[utf8]{inputenc}
\usepackage[ruled, lined, longend, linesnumbered]{algorithm2e}
\usepackage{amsmath,multicol}
\usepackage{subcaption,textcomp,comment}
\usepackage{url}
\usepackage[most,breakable,skins]{tcolorbox} 

\usepackage{rotating}
\usepackage{authblk}
\usepackage[symbol]{footmisc}

\usepackage{booktabs}
\usepackage{multirow}
\usepackage{enumitem}
\usepackage{amssymb}
\usepackage{graphicx}

\usepackage{tabularx}
\usepackage{array}

\usepackage{rotating}
\newcolumntype{L}[1]{>{\raggedright\let\newline\\\arraybackslash\hspace{0pt}}m{#1}}
\newcolumntype{C}[1]{>{\centering\let\newline\\\arraybackslash\hspace{0pt}}m{#1}}
\newcolumntype{R}[1]{>{\raggedleft\let\newline\\\arraybackslash\hspace{0pt}}m{#1}}

\DeclareSymbolFont{extraup}{U}{zavm}{m}{n}
\DeclareMathSymbol{\varheart}{\mathalpha}{extraup}{86}
\DeclareMathSymbol{\vardiamond}{\mathalpha}{extraup}{87}

\usepackage{tikz}
\usetikzlibrary{external}
\usepackage[edges]{forest}
\definecolor{hidden-draw}{RGB}{20,68,106}
\definecolor{hidden-pink}{RGB}{255,245,247}
\definecolor{hidden-red}{RGB}{180,0,0}

\newtcolorbox{remark}{
  colback=blue!5!white, 
  colframe=blue!75!black, 
  boxrule=0pt, 
  leftrule=2pt, 
  rightrule=2pt, 
  boxsep=5pt, 
  arc=0pt, 
  left=5pt, 
  right=5pt, 
  top=0pt, 
  bottom=0pt
}

\title{Small Language Models:\\Survey, Measurements, and Insights}

\author{
\bf Zhenyan Lu$^{\clubsuit}$$^\diamondsuit$\footnote[2], , Xiang Li$^{\clubsuit}$\footnote[2], , Dongqi Cai$^{\clubsuit}$, Rongjie Yi$^{\clubsuit}$, Fangming Liu$^\diamondsuit$, Xiwen Zhang$^\varheart$,\\
\vspace{-0.1cm}
\bf Nicholas D. Lane$^\heartsuit$, Mengwei Xu$^{\clubsuit}$\\
\vspace{0.3cm}
$^{^\clubsuit}$Beijing University of Posts and Telecommunications (BUPT)\\
$^\diamondsuit$Peng Cheng Laboratory\\
$^\varheart$Helixon Research\\
$^\heartsuit$University of Cambridge\\
\vspace{0.3cm}
Website: \url{https://github.com/UbiquitousLearning/SLM_Survey}
}

\date{July 2024}

\begin{document}

\maketitle
\renewcommand{\thefootnote}{\fnsymbol{footnote}} 
\footnotetext[2]{Zhenyan Lu and Xiang Li contributed equally to this work.} 

\input{sec-abstract}
	





\input{sec-intro}
\input{sec-overview}
\input{sec-capability}
\input{sec-cost}
\input{sec-conclusion}

\bibliographystyle{plain}
\bibliography{ref}

\end{document}

%% file: sec-abstract.tex
\begin{abstract}

    Small language models (SLMs), despite their widespread adoption in modern smart devices, have received significantly less academic attention compared to their large language model (LLM) counterparts, which are predominantly deployed in data centers and cloud environments. While researchers continue to improve the capabilities of LLMs in the pursuit of artificial general intelligence, SLM research aims to make machine intelligence more accessible, affordable, and efficient for everyday tasks. Focusing on transformer-based, decoder-only language models with 100M–5B parameters, we survey 70 state-of-the-art open-source SLMs, analyzing their technical innovations across three axes: architectures, training datasets, and training algorithms. In addition, we evaluate their capabilities in various domains, including commonsense reasoning, mathematics, in-context learning, and long context. To gain further insight into their on-device runtime costs, we benchmark their inference latency and memory footprints. Through in-depth analysis of our benchmarking data, we offer valuable insights to advance research in this field.

    
\end{abstract}

%% file: sec-intro.tex
\section{Introduction}

The evolution of language models is diverging.
On one hand, in the pursuit of artificial general intelligence (AGI) and following the scaling law, increasingly large language models (LLM) have been born in datacenters that host hundreds of thousands of GPUs~\cite{kaplan2020scaling,xu2024survey}.
The aim of this path is to demonstrate that machines can solve the most challenging language tasks, with the ultimate goal of advancing human civilization by pushing the boundaries of science and technology.
On the other hand, there is a growing focus on \textbf{small language models (SLM)}, designed for resource-efficient deployment on personal devices such as desktops, smartphones, and even wearables. 
The vision behind SLMs is to democratize access to machine intelligence, making it both accessible and affordable to people everywhere.
This approach seeks to make intelligence ubiquitous and practical, available to anyone, anywhere, at any time -- much like the human brain, which everyone possesses.

Both LLM and SLM are important in reshaping our daily lives, yet the latter receives significantly less attention in academia.
There has been very limited literature exploring SLM capabilities~\cite{lepagnol2024small,schick2020s,zhou2023mini} or their runtime cost on devices~\cite{li2024large,laskaridismobile,xu2024device}, often with limited scale or depth.
In the real world, however, SLMs have already been integrated into commercial off-the-shelf (COTS) devices on a massive scale~\cite{yuan2023mobile,dubiel2024device}.
For instance, the latest Google/Samsung smartphones have built-in LLM services (Gemini Nano), allowing third-party mobile apps to leverage LLM capabilities through prompts and LoRA modules~\cite{google-aicore}.
The most recent iOS system on iPhones and iPads also includes an on-device foundation model, deeply integrated with the operating system for better performance and privacy~\cite{apple-intelligence}.
Beyond resource-constrained scenarios, SLMs also excel superior performance in certain domain-specific tasks~\cite{li2024small}.

This work presents the first comprehensive survey of SLMs, thoroughly discussing their capabilities, runtime costs, and innovations in recent years.
The scope of this survey is limited to those \textit{language models with 100M--5B parameters in decoder-only transformer architecture}\footnote{The definition of ``small'' could drift over time, considering that device memory is increasing over time and can host larger ``small language models'' in the future. 
We set 5B as the upper limit for the size of SLMs, since as of Seq. 2024, 7B LLMs are still mostly deployed in the cloud.}, which covers the range of devices from low-end IoT/wearable gadgets like smartwatches to high-end mobile devices such as smartphones and tablets.
In total, we collected 57 popular SLMs and fully benchmarked their capabilities (commonsense reasoning, math, in-context learning, long-context retrieval, etc.) and on-device runtime costs (prefill and decode speed, memory footprint, energy consumption, etc.).

With the insights from comprehensive benchmarking and thorough investigation, We try to answer the following questions concerning SLMs: 
``What is the evolving path of SLMs?''
``What datasets or training strategies are more likely to produce a highly capable SLM?''
``How different SLM architecture (e.g., depth, width, atten type) and the deployment environments (quantization algorithms, hardware type, etc) impact runtime performance?''
We expect the work to reveal an all-sided landscape of SLM and benefit the research community, including those working on the algorithm, model, system, and hardware levels.



In summary, we make the following contributions in this work.

\begin{itemize}
    \item We exhaustively review the small language models released in recent years, summarize their key innovations, and benchmark their capability as well as on-device cost.
    \item Through such in-depth investigation, we obtain valuable insights from open-sourced SLMs, which can potentially benefit future SLM research. We also summarize a few potential research topics in SLM.
    \item We make all results and benchmark tools public to advance and facilitate the SLM research.
\end{itemize}

%% file: sec-overview.tex
\section{SLM Architecture, Datasets, and Training}

\input{sec-overview-slm.tex}

\input{sec-innovation-architecture.tex}

\input{sec-innovation-dataset.tex}

\input{sec-innovation-algorithm.tex}

%% file: sec-overview-slm.tex
\subsection{Overview of SLMs}

\input{fig-model-timeline.tex}

\input{tab-model-details.tex}

As shown in Figure~\ref{fig:model-timeline}, SLMs have gained increasing attention from both the research and industrial communities.
Notably, since the end of 2023, the number of SLM models has surged significantly. 
To understand their capability and cost, we comprehensively collect SLMs based on the following criteria:
(1) We only collect models with decoder-only transformer architecture (which can be traced to GPT-2), for their superior performance and real-world deployment.
Currently we do not include variants of transformers such as RWKV~\cite{peng2023rwkv} and Mamba~\cite{gu2023mamba}.
(2) We only collect SLMs with open weights, so that we can evaluate them.
(3) The weight range of SLMs in this work is defined between 100M to 5B.
(4) The survey mainly focuses on the base knowledge obtained from pre-training process, thereby we only collect the base models of SLMs, except those provided only the instruct versions (Microsoft Phi and StabilityAI StableLM).
We also exclude models that are fine-tuned on other pre-trained models.

With such criteria, we select 70 SLMs as detailed in Table~\ref{tab:model-details}. 
Our selection encompasses a wide range of models from both industry and academia, based on factors such as model architecture, parameter size, and data availability. 
While all selected SLMs share similar architectures, they differ in specific hyperparameters and training datasets, with some datasets remaining closed-source. 
These variations lead to differing performance across tasks, as we will discuss in the following sections.



%% file: fig-model-timeline.tex

\begin{figure*}[h]
	\centering
	\includegraphics[width=0.98\textwidth,height=0.29\textwidth]{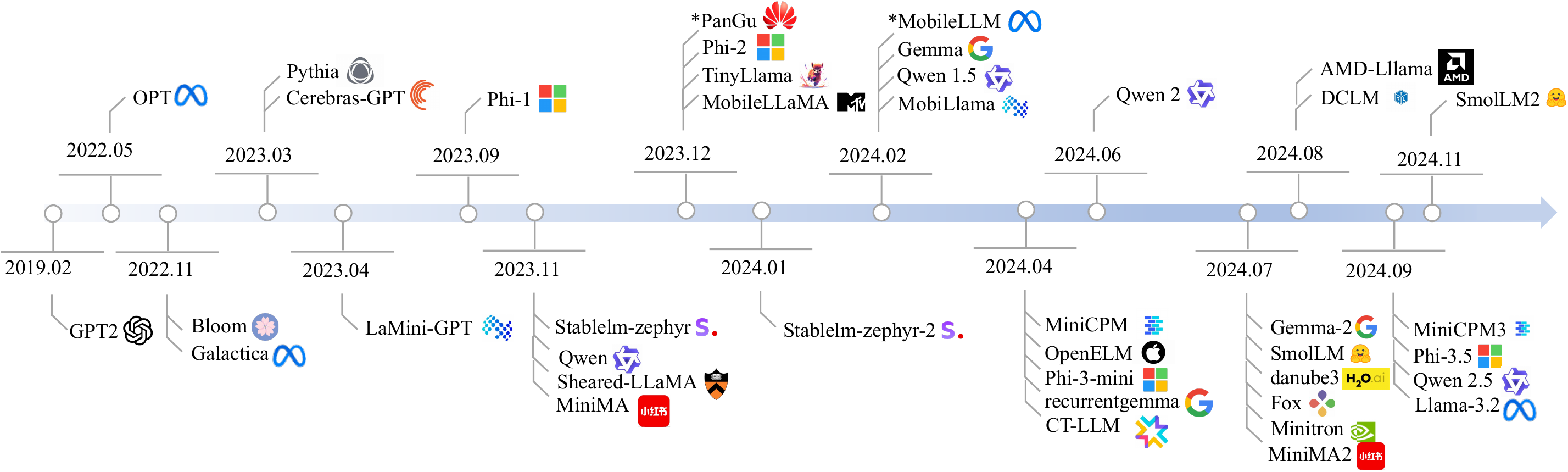}
	\caption{
        An overview of SLMs. * indicates the models are not open-sourced so will not be benchmarked. 
		}
	\label{fig:model-timeline}
\end{figure*}

%% file: tab-model-details.tex
\begin{table}[]
    \vspace{-30pt}
    \centering
    \huge
    \resizebox{\columnwidth}{!}{%
    \begin{tabular}{c|c|c|c|c|c|c|c|c|c|c|c}
    \hline
    \textbf{Affiliation} & \textbf{\begin{tabular}[c]{@{}c@{}}Model\\ name\end{tabular}} & \textbf{Size} & \textbf{Date} & \textbf{Attention} & \textbf{\begin{tabular}[c]{@{}c@{}}Layer\\ number\end{tabular}} & \textbf{\begin{tabular}[c]{@{}c@{}}Hidden\\ size\end{tabular}} & \textbf{\begin{tabular}[c]{@{}c@{}}Head \\ num\end{tabular}} & \textbf{Activation} & \textbf{\begin{tabular}[c]{@{}c@{}}Vocab.\\ size\end{tabular}} & \textbf{\begin{tabular}[c]{@{}c@{}}Open \\ training \\ datasets\end{tabular}} & \textbf{\begin{tabular}[c]{@{}c@{}}Max\\ context\\ window\end{tabular}} \\ \hline
    \multirow{8}{*}{Meta~\centering \raisebox{-0.2\height}{\includegraphics[width=1cm]{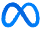}}} & \multirow{4}{*}{OPT~\cite{facebook_opt_125m}} & 125M & \multirow{4}{*}{2022.05} & \multirow{4}{*}{MHA} & 12 & 768 & 12 & \multirow{4}{*}{ReLU} & \multirow{4}{*}{50k} & \multirow{4}{*}{$\checkmark$} & \multirow{4}{*}{2k} \\
    & & 350M &  &  & 24 & 1024 & 16 &  &  &  &  \\
    & & 1.3B &  &  & 24 & 2048 & 32 &  &  &  &  \\
    & & 2.7B &  &  & 32 & 2560 & 32 &  &  &  &  \\ \cline{2-12}
    & \multirow{2}{*}{Galactica~\cite{facebook_galactica_125m}} & 125M & \multirow{2}{*}{2022.11} & \multirow{2}{*}{MHA} & 12 & 768 & 12 & \multirow{2}{*}{GELU} & \multirow{2}{*}{50k} & \multirow{2}{*}{  } & \multirow{2}{*}{2k} \\
    & & 1.3B &  &  & 24 & 2048 & 32 &  &  &  &  \\ \cline{2-12}
    & \multirow{2}{*}{Llama-3.2~\cite{meta_llama_3_2}} & 3B & 2024.09 & GQA & 28 & 3072 & 24 & SiLU & 128k &  & 131k \\
    & & 1B & 2024.09 & GQA & 16 & 2048 & 32 & SiLU & 128k &  & 131k \\ \hline
    \multirow{4}{*}{BigScience~\centering \raisebox{-0.2\height}{\includegraphics[width=1cm]{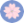}}} & \multirow{2}{*}{Bloom~\cite{bigscience_bloom_560m}} & 560M & \multirow{2}{*}{2022.11} & \multirow{2}{*}{MHA} & \multirow{4}{*}{24} & 1024 & \multirow{4}{*}{16} & \multirow{4}{*}{$\text{GELU}_{\text{tanh}}$} & \multirow{4}{*}{251k} & \multirow{4}{*}{$\checkmark$} & \multirow{4}{*}{2k} \\
    & & 1.1B &  &  &  & 1536 &  &  &  &  &  \\ \cline{2-5} \cline{7-7}
    & \multirow{2}{*}{Bloomz~\cite{bigscience_bloomz_1b1}} & 1.1B & \multirow{2}{*}{2022.11} & \multirow{2}{*}{MHA} &  & 1536 &  &  &  &  &  \\
    & & 560M &  &  &  & 1024 &  &  &  &  &  \\ \hline
    \multirow{5}{*}{EleutherAI\centering \raisebox{-0.2\height}{\includegraphics[width=1.4cm]{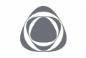}}} & \multirow{5}{*}{Pythia~\cite{eleutherai_pythia_410m}} & 160M & \multirow{5}{*}{2023.03} & \multirow{5}{*}{MHA} & 12 & 768 & 12 & \multirow{5}{*}{GELU} & \multirow{5}{*}{50k} & \multirow{5}{*}{$\checkmark$} & \multirow{5}{*}{2k} \\
    & & 410M &  &  & 24 & 1024 & 16 &  &  &  &  \\
    & & 1B &  &  & 16 & 2048 & 8 &  &  &  &  \\
    & & 1.4B &  &  & 24 & 2048 & 16 &  &  &  &  \\
    & & 2.8B &  &  & 32 & 2560 & 32 &  &  &  &  \\ \hline
    \multirow{5}{*}{Cerebras~\centering \raisebox{-0.2\height}{\includegraphics[width=1cm]{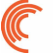}}} & \multirow{5}{*}{Cerebras-GPT~\cite{cerebras_cerebras_gpt_111m}} & 111M & \multirow{5}{*}{2023.03} & \multirow{5}{*}{MHA} & 10 & 768 & 12 & \multirow{5}{*}{GELU} & \multirow{5}{*}{50k} & \multirow{5}{*}{$\checkmark$} & \multirow{5}{*}{2k} \\
    & & 256M &  &  & 14 & 1088 & 17 &  &  &  &  \\
    & & 590M &  &  & 18 & 1536 & 12 &  &  &  &  \\
    & & 1.3B &  &  & 24 & 2048 & 16 &  &  &  &  \\
    & & 2.7B &  &  & 32 & 2560 & 32 &  &  &  &  \\ \hline
    \multirow{4}{*}{Microsoft~\centering \raisebox{-0.2\height}{\includegraphics[width=1cm]{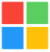}}} & Phi-1~\cite{microsoft_phi} & 1.3B & 2023.09 & MHA & 24 & 2048 & 32 & $\text{GELU}_{\text{tanh}}$ & 51k &    & \multirow{2}{*}{2k} \\ \cline{2-11}
    & Phi-1.5~\cite{microsoft_phi_1_5} & 1.3B & 2023.09 & MHA & 24 & 2048 & 32 & $\text{GELU}_{\text{tanh}}$ & 51k &    &  \\ \cline{2-12} 
    & Phi-2~\cite{microsoft_phi_2} & 2.7B & 2023.12 & MHA & 32 & 2560 & 32 & $\text{GELU}_{\text{tanh}}$ & 51k &    & 2k \\ \cline{2-12} 
    & Phi-3-mini*~\cite{microsoft_phi_3_mini} & 3.8B & 2024.04 & MHA & 32 & 3072 & 32 & SiLU & 32k &    & 4k \\ \cline{2-12}
    & Phi-3.5-mini* & 2.7B & 2024.09 & MHA & 32 & 3072 & 32 & SiLU & 32k &    & 4k \\ \hline
    \multirow{2}{*}{StabilityAI\centering \raisebox{-0.1\height}{~\includegraphics[width=1cm]{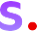}}} & StableLM-zephyr*~\cite{stabilityai_stablelm_zephyr_3b} & 3B & 2023.11 & MHA & 32 & 2560 & 32 & SiLU & 50k & $\checkmark$ & 1k \\ \cline{2-12} 
    & StableLM-2-zephyr*~\cite{stabilityai_stablelm_2_zephyr} & 1.6B & 2024.01 & MHA & 24 & 2048 & 32 & SiLU & 100k & $\checkmark$ & 4k \\ \hline
    TinyLlama~\includegraphics[width=1cm]{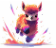} & TinyLlama~\cite{tinyllama} & 1.1B & 2023.12 & GQA & 22 & 2048 & 32 & SiLU & 32k & $\checkmark$ & 2k \\ \hline
    Meituan~\centering \raisebox{-0.25\height}{\includegraphics[width=0.9cm]{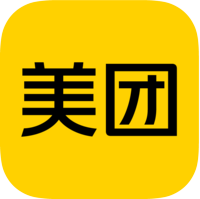}} & MobileLLaMA~\cite{mobilellama} & 1.4B & 2023.12 & GQA & 24 & 2048 & 16 & SiLU & 32k & $\checkmark$ & 2k \\ \hline
    \multirow{7}{*}{Alibaba~\centering \raisebox{-0.2\height}{\includegraphics[width=1cm]{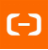}}} & Qwen 1~\cite{qwen_1} & 1.8B & 2023.11 & MHA & 24 & 2048 & 16 & SiLU & 152k &    & 8k \\ \cline{2-12} 
    & Qwen 1.5~\cite{qwen_1_5} & 0.5B & 2024.02 & MHA & 24 & 1024 & 16 & SiLU & 152k &    & 32k \\ \cline{2-12} 
    & \multirow{2}{*}{Qwen 2~\cite{qwen_2}} & 1.8B & \multirow{2}{*}{2024.06} & \multirow{2}{*}{MHA} & 24 & 2048 & 16 & \multirow{2}{*}{SiLU} & \multirow{2}{*}{152k} &  & 32k \\
    & & 4B &  &  & 40 & 2560 & 20 &  &  &  & 32k \\ \cline{2-12} 
    & \multirow{3}{*}{Qwen 2.5~\cite{qwen_2.5}} & 0.5B & \multirow{3}{*}{2024.09} & \multirow{3}{*}{GQA} & 24 & 896 & 14 & \multirow{3}{*}{SiLU} & \multirow{3}{*}{152k} &  & \multirow{3}{*}{32k} \\
    & & 1.5B &  &  & 28 & 1536 & 12 &  &  &  &  \\
    & & 3B &  &  & 36 & 2048 & 16 &  &  &  &  \\ \hline
    \multirow{4}{*}{MBZUAI~\includegraphics[width=1cm]{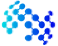}} & \multirow{2}{*}{MobiLlama~\cite{mobillama}} & 0.5B & \multirow{2}{*}{2024.02} & \multirow{2}{*}{GQA} & \multirow{2}{*}{22} & \multirow{2}{*}{2048} & \multirow{2}{*}{32} & \multirow{2}{*}{SiLU} & \multirow{2}{*}{32k} & \multirow{2}{*}{$\checkmark$} & \multirow{2}{*}{2k} \\
    & & 1B &  &  &  &  &  &  &  &  &  \\ \cline{2-12} 
    & \multirow{2}{*}{LaMini-GPT~\cite{mbzuai_lamini_gpt_774m}} & 774M & \multirow{2}{*}{2023.04} & \multirow{2}{*}{MHA} & 36 & 1280 & 20 & \multirow{2}{*}{$\text{GELU}_{\text{tanh}}$} & \multirow{2}{*}{50k} & \multirow{2}{*}{  } & \multirow{2}{*}{1k} \\
    & & 1.5B &  &  & 48 & 1600 & 25 &  &  &  &  \\ \hline
    \multirow{3}{*}{Google\centering \raisebox{-0.1\height}{\includegraphics[width=1.4cm]{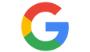}}} & Gemma~\cite{gemma} & 2B & 2024.02 & MQA & 18 & 2048 & 8 & GELU & 256k &    & 8k \\ \cline{2-12} 
    & recurrentGemma~\cite{recurrentgemma} & 2B & 2024.04 & MQA & 26 & 2560 & 10 & $\text{GELU}_{\text{tanh}}$ & 256k &    & 8k \\ \cline{2-12} 
    & Gemma-2 & 2B & 2024.07 & GQA & 26 & 2304 & 8 & $\text{GELU}_{\text{tanh}}$ & 256k &    & 8k \\ \hline
    \multirow{3}{*}{OpenBMB\centering \raisebox{-0.2\height}{\includegraphics[width=1.2cm]{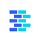}}} & \multirow{2}{*}{MiniCPM~\cite{minicpm}} & 1B & \multirow{2}{*}{2024.04} & \multirow{2}{*}{GQA} & 52 & 1536 & 24 & \multirow{2}{*}{SiLU} & 73k & \multirow{2}{*}{  } & 128k \\
    & & 2B &  &  & 40 & 2304 & 36 &  & 123k &  & 131k \\ \cline{2-12} 
    & MiniCPM3~\cite{minicpm3} & 4B & 2024.09 & MLA &  62 & 2560 & 40 & SiLU & 73k &  &  \\ \hline
    \multirow{4}{*}{Apple\centering \raisebox{-0.2\height}{\includegraphics[width=1.4cm]{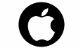}}} & \multirow{4}{*}{OpenELM~\cite{openelm}} & 270M & \multirow{4}{*}{2024.04} & \multirow{4}{*}{GQA} & 16 & 1280 & 12-20 & \multirow{4}{*}{SiLU} & \multirow{4}{*}{32k} & \multirow{4}{*}{$\checkmark$} & \multirow{4}{*}{2k} \\
    & & 450M &  &  & 20 & 1536 & 12-24 &  &  &  &  \\
    & & 1.1B &  &  & 28 & 2048 & 16-32 &  &  &  &  \\
    & & 3B &  &  & 36 & 3072 & 12-24 &  &  &  &  \\ \hline
    \multirow{2}{*}{H2O\centering \raisebox{-0.1\height}{~\includegraphics[width=1.6cm]{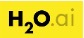}}} &\multirow{2}{*}{danube3~\cite{h2oai_danube}} & 0.5B & \multirow{2}{*}{2024.07} &  \multirow{2}{*}{GQA} & 16 & 1536 & 16 & SiLU & 32k &  & 8k \\
    &  & 4B &  &  & 24 & 3840 & 32 & SiLU & 32k &  & 8k  \\ \hline
    TensorOpera AI\centering \raisebox{-0.3\height}{~\includegraphics[width=0.8cm]{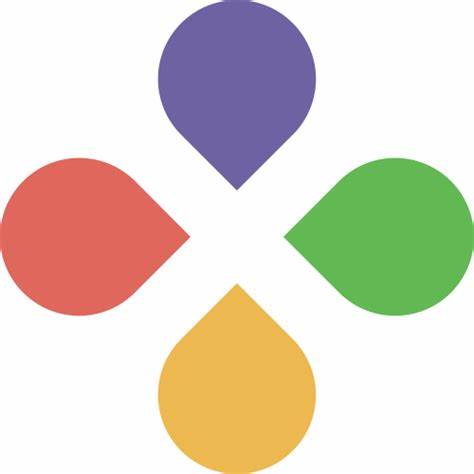}} & Fox~\cite{tensoropera_fox} & 1.6B & 2024.07 & GQA &  32 & 2048 & 16 & SiLU & 32k &  & 8k \\ \hline
    \multirow{5}{*}{HuggingFace~\centering \raisebox{-0.2\height}{\includegraphics[width=1cm]{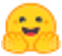}}} & \multirow{3}{*}{SmolLM~\cite{smollm}} & 135M & \multirow{3}{*}{2024.07} & GQA & 30 & 576 & 9 & \multirow{3}{*}{SiLU} & \multirow{3}{*}{49k} & \multirow{3}{*}{$\checkmark$} & \multirow{3}{*}{2k} \\
    & & 360M &  & GQA & 32 & 960 & 15 &  &  &  &  \\
    & & 1.7B &  & MHA & 24 & 2048 & 32 &  &  &  &  \\ \cline{2-12} 
    &   \multirow{2}{*}{SmolLM2~\cite{smollm2}} & 360M & \multirow{2}{*}{2024.11} & GQA & 30 & 576 & 9 &  &  &  &  \\
    & & 1.7B &  & MHA & 24 & 2048 & 32 &  &  &  &  \\ \hline
    Toyota\centering \raisebox{-0.3\height}{\includegraphics[width=1.6cm]{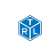}} & DCLM~\cite{dclm} & 1.4B & 2024.08 & MHA &  24 & 2048 & 16 & SiLU & 50k &
 $\checkmark$ & 50k \\ \hline
    DataBricks~\centering \raisebox{-0.2\height}{\includegraphics[width=1cm]{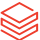}} & Dolly-v2*~\cite{databricks_Dolly_v2_3b} & 3B & 2023.04 & MHA & 32 & 2560 & 32 & GELU & 50k &    & 2k \\ \hline
    AllenAI~\centering \raisebox{-0.2\height}{\includegraphics[width=1cm]{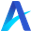}} & OLMo~\cite{allenai_olmo_1b_hf} & 1.18B & 2024.04 & MHA & 16 & 2048 & 16 & SiLU & 50k & $\checkmark$ & 50k \\ \hline
    \multirow{2}{*}{Princeton~\centering \raisebox{-0.2\height}{\includegraphics[width=1cm]{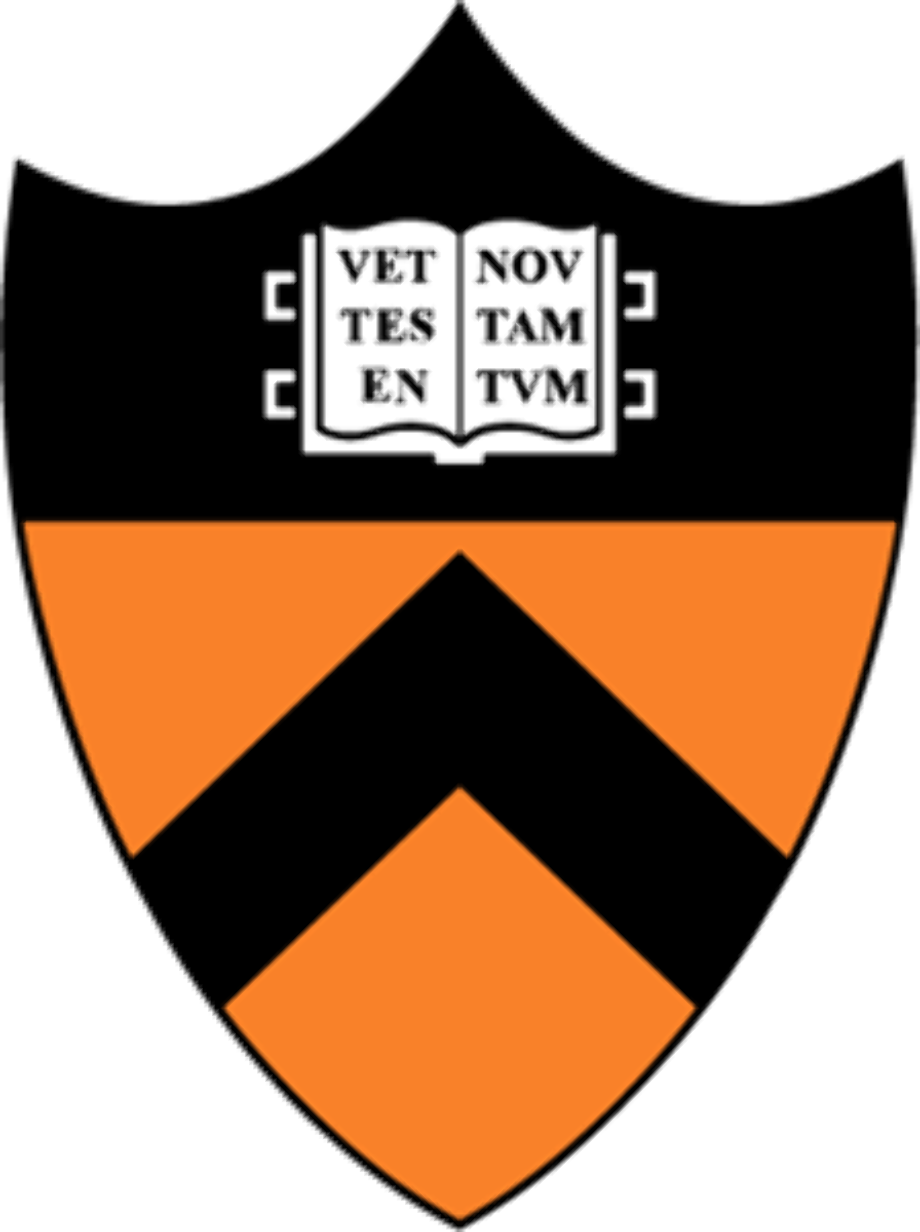}}} & \multirow{2}{*}{Sheared-LLaMA~\cite{princeton_sheared_llama}} & 1.3B & 2023.11 & MHA & 24 & 2048 & 16 & SiLU & 32k &  & 4096 \\
   & & 2.7B & 2023.11 & MHA & 32 & 2560 & 20 & SiLU & 32k &  & 4096 \\ \hline
   \multirow{2}{*}{Xiaohongshu~\centering \raisebox{-0.2\height}{\includegraphics[width=1cm]{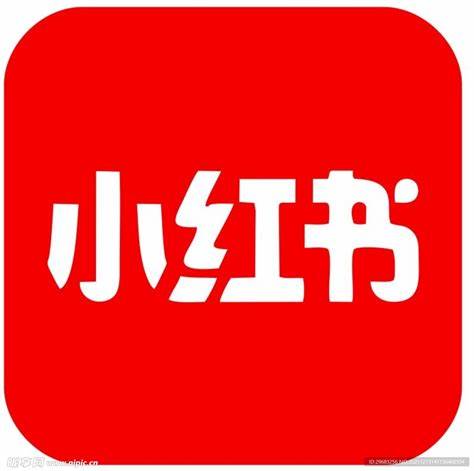}}} & MiniMA~\cite{xiaohongshu_minima} & 3B & 2023.11 & \multirow{2}{*}{UKN} & 24 & 3072 & 24 & SiLU & 49k &  & 4096 \\ 
    & MiniMA2~\cite{xiaohongshu_minima-2} & 1B & 2024.07 &  & 18 & 2304 & 18 & SiLU & 49k &  & 4096 \\ \hline
    \multirow{1}{*}{Nvidia~\centering \raisebox{-0.2\height}{\includegraphics[width=1cm]{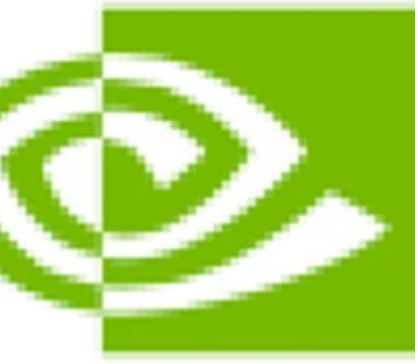}}} & Minitron~\cite{nvidia_minitron} & 4B & 2024.07 & GQA & 32 & 3072 & 24 & ReLU2 & 256k &  & 4096 \\ \hline
    \multirow{1}{*}{M.A.P.~\centering \raisebox{-0.2\height}{\includegraphics[width=0.8cm]{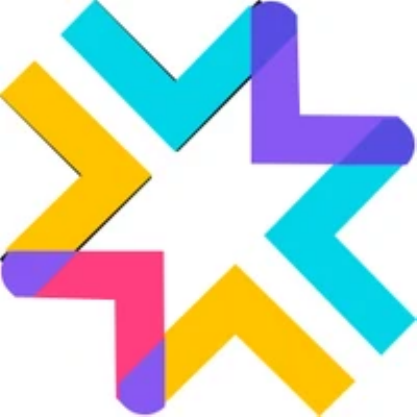}}} & CT-LLM~\cite{map_ct_llm} & 2B & 2024.04 & MHA & 32 & 2048 & 16 & SiLU & 125k &  & 4096 \\ \hline
    \multirow{1}{*}{AMD~\centering \raisebox{-0.2\height}{\includegraphics[width=0.8cm]{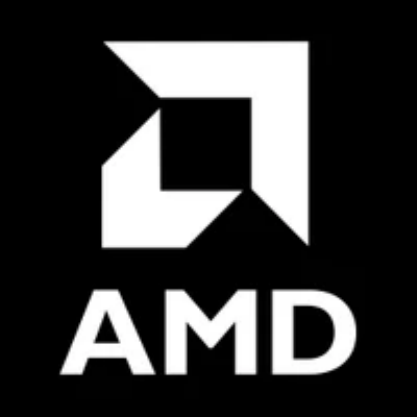}}} & AMD-Llama~\cite{amd_llama} & 135M & 2024.08 & MHA & 12 & 768 & 12 & SiLU & 32k &  & 2048 \\ \hline
    \end{tabular}%
    }
    \caption{Detailed configurations of SLMs benchmarked. We mainly use the base models in experiments, with exceptions of StableLM, Phi-3/3.5 and Dolly-v2 (marked with *) that only provide the instruct version. 
    }
    \label{tab:model-details}
\end{table}

%% file: sec-innovation-architecture.tex
\subsection{Model Architecture}



While we focus on only decoder-only transformer SLMs, their specific configurations still diversify, as shown in Figure \ref{fig:innovation-model-arch}(a).
The core of Transformer is the multi-head self-attention(MHA) mechanism and the Feed-Forward Neural Network(FFN).

\input{figs-innovation/fig-innovation-model-arch}

\textbf{Model architecture analysis.}
We conduct statistical analysis on the following several components of the model architecture: 1) The type of self-attention; 2) The type of feed-forward neural network; 3) The intermediate ratio of the feed-forward network; 4) The activation function of the feed-forward neural network; 5) The type of layer normalization; 6) The vocabulary size.
Figure \ref{fig:innovation-model-arch}(a) shows the architecture of SLM and the pie chart shows the distribution of six components.
Figure \ref{fig:innovation-model-arch}(b) shows how these distributions change over time.

1) \textit{The type of self-attention.} 
The self-attention mechanism is the core of the Transformer model. 
In general, SLMs mainly use three types of attention mechanism: Multi-Head Attention (MHA), Multi-Query Attention (MQA), Group-Query Attention (GQA) and Multi-Head Latent Attention(MLA). 
Multi-Head Attention is a mechanism that allows the model to focus on different parts of the input data simultaneously by employing multiple attention heads, which is the most widely used self-attention mechanism in the Transformer models. 
Multi-Query Attention simplifies multi-head attention by using a single shared query across all heads but allowing different key and value projections. This reduces the complexity in both space and time.
Group-Query Attention is a variant of multi-head attention that reduces computational complexity by sharing query representations across multiple heads, while allowing separate key and value representations. The idea is to use fewer query groups but still preserve a level of diversity in the attention mechanism.
Multi-Head Latent Attention achieves better results than MHA through low-rank key-value joint compression, and requires much less Key-Value(KV) Cache. 

Figure \ref{fig:innovation-model-arch}(b)\normalsize{\textcircled{\scriptsize{1}}}\normalsize\enspace shows the changing situation of choosing three self-attention mechanisms during these time periods from 2022 to 2024. We can see that MHA is gradually being phased out and replaced by GQA.

2) \textit{The type of feed-forward neural network.}
Feed-forward network can be summarized into two types: the Standard FFN and the Gated FFN. 
The Standard FFN is a two-layer neural network with a activation function. The Gated FFN adds an additional gate layer. 

The Figure \ref{fig:innovation-model-arch}(b)\normalsize{\textcircled{\scriptsize{2}}}\normalsize\enspace shows the changing situation of type of FFN during these time periods from 2022 to 2024. It shows that Standard FFN is gradually being phased out and replaced by Gated FFN.

3) \textit{The intermediate ratio of the feed-forward neural network.} 
The intermediate ratio of FFN is the ratio of the intermediate dimension to the hidden dimension. 
Figure \ref{fig:innovation-model-arch}(b)\normalsize{\textcircled{\scriptsize{3}}}\normalsize\enspace shows that the intermediate ratio of the Standard FFN is commonly set to be 4, while the intermediate ratio of the Gated FFN is rather diversified ranging from 2 to 8. 

4) \textit{The activation function of the feed-forward neural network.} 
There are 4 main kinds of activation functions used in FFN: ReLU (Rectified Linear Unit), GELU (Gaussian Error Linear Unit), $\text{GELU}_{\text{tanh}}$, SiLU (Sigmoid Linear Unit). 
Observed from Figure \ref{fig:innovation-model-arch}(b)\normalsize{\textcircled{\scriptsize{4}}}\normalsize\enspace, the activation function of FFN was mostly ReLU in 2022, and then changed to GELU and its variants in 2023.
For those released in 2024, SiLU becomes the dominant type. 

5) \textit{The type of layer normalization.} 
There are two main types of layer normalization: LayerNorm and RMSNorm. 
The Figure \ref{fig:innovation-model-arch}(b)\normalsize{\textcircled{\scriptsize{5}}}\normalsize\enspace shows the changing situation of type of the type of layer normalization during these time periods from 2022 to 2024. 
layer normalization is gradually being phased out and replaced by RMS normalization.

6) \textit{The vocabulary size.}
The vocabulary size is the total number of unique tokens that an SLM can recognize.
The Figure \ref{fig:innovation-model-arch}(b)\normalsize{\textcircled{\scriptsize{6}}}\normalsize\enspace shows the changing situation of the vocabulary size during these time periods from 2022 to 2024. We can see that the vocabulary size of the model is gradually increasing.
The vocabulary of the latest models is often larger than 50k  

\textbf{Model architecture innovations.}
While the vanilla transformer architecture has been well recognized for its scaling ability, there still exist a few architecture-level innovations in the tested SLMs, namely parameter sharing and layer-wise parameter scaling. 

1) Parameter Sharing.
Parameter Sharing is a technique used in large language models to reuse the same set of weights across different layers or components of the network. This approach allows the model to significantly reduce the number of parameters, leading to more efficient training and inference, while maintaining performance.

\textit{Embedding-lm\_head sharing.} 
Sharing the weights of the embedding with the final lm\_head layer is the most common weight sharing technique.  
It is the sharing of the word embedding layer and has nothing to do with the rotary position encoding. 
Models such as Gemma, and Qwen all used this sharing technique. 

\textit{layer-wise attention/FFN sharing.}
In this approach, the same set of weights is reused across multiple layers of the model. 
This is commonly seen in SLM/LLM, where all the transformer layers share the same parameters.
For example, MobiLLaMA shares the weights of the FFN of all the transformer blocks;
MobileLLM shares the weights of the Attention and FFN of two adjacent transformer blocks.


2) Layer-wise parameter scaling.
This technique was proposed and used by OpenELM.
Traditional SLMs use the same configuration for each transformer layer in the model, resulting in a uniform allocation of parameters across layers.
Unlike these models, each transformer layer in OpenELM has a different configuration (e.g., number of heads and feed forward network dimension), resulting in variable number of parameters in each layer of the model. This lets OpenELM to better utilize the available parameter budget for achieving higher accuracies. 

3) Nonlinearity compensation. 
PanGu-$\pi$ analyzes the state-of-the-art language model architectures and observes the feature collapse problem. 
PanGu-$\pi$ adopts two techniques for nonlinearity compensation of language model to solve the feature collapse problem. 
The series activation function is adapted to FFN, and the augmented shortcuts are integrated into MHA, which effectively introduces more nonlinearity into the Transformer architecture.

\begin{remark}
\textbf{Insights}: We make two key observations in SLM architectures.
\begin{itemize}[leftmargin=*]
    \item As of August 2024, a typical SLM architecture tends to use group-query attention, gated FFN with SiLU activation, an intermediate ratio of FFN between 2 and 8, RMS normalization, and a vocabulary size larger than 50K.
    However, the choice of such settings is mostly empirical, without strict and public validation on the superiority of such model's capacity.
    Instead, the architecture innovations have relative larger impacts on the runtime performance on devices, as will be shown in $\S$\ref{sec:cost}.
    \item The innovations to the transformer architecture is limited in nowaday SLMs.
    For the few that did contribute architectural innovation (except embedding-lm head sharing), we do not observe strong evidence showing them being significantly superior to the vanilla transformer, and neither are them being generally adopted or studied across different research groups or companies.
    The significance of those innovations remain to be explored and validated.
\end{itemize}
\end{remark}

%% file: figs-innovation/fig-innovation-model-arch.tex
\begin{figure*}[htbp]
    
    \centering
    \begin{minipage}[b]{\textwidth}
        \includegraphics[width=0.99\textwidth]{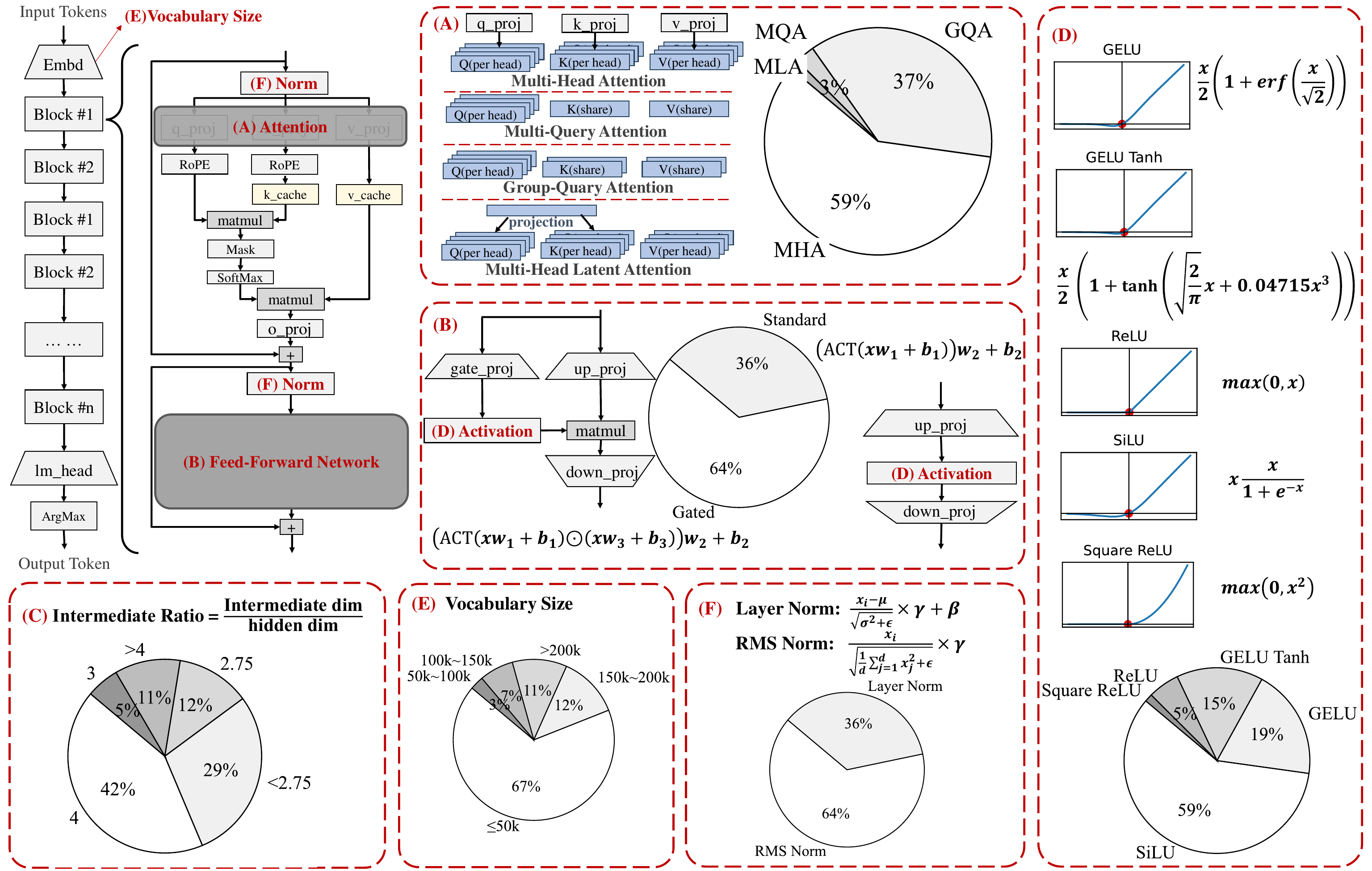}
        \subcaption{The architecture. 
        }
    \end{minipage}
    
    \begin{minipage}[b]{\textwidth}
        \vspace{0pt} 
        \includegraphics[width=0.99\textwidth]{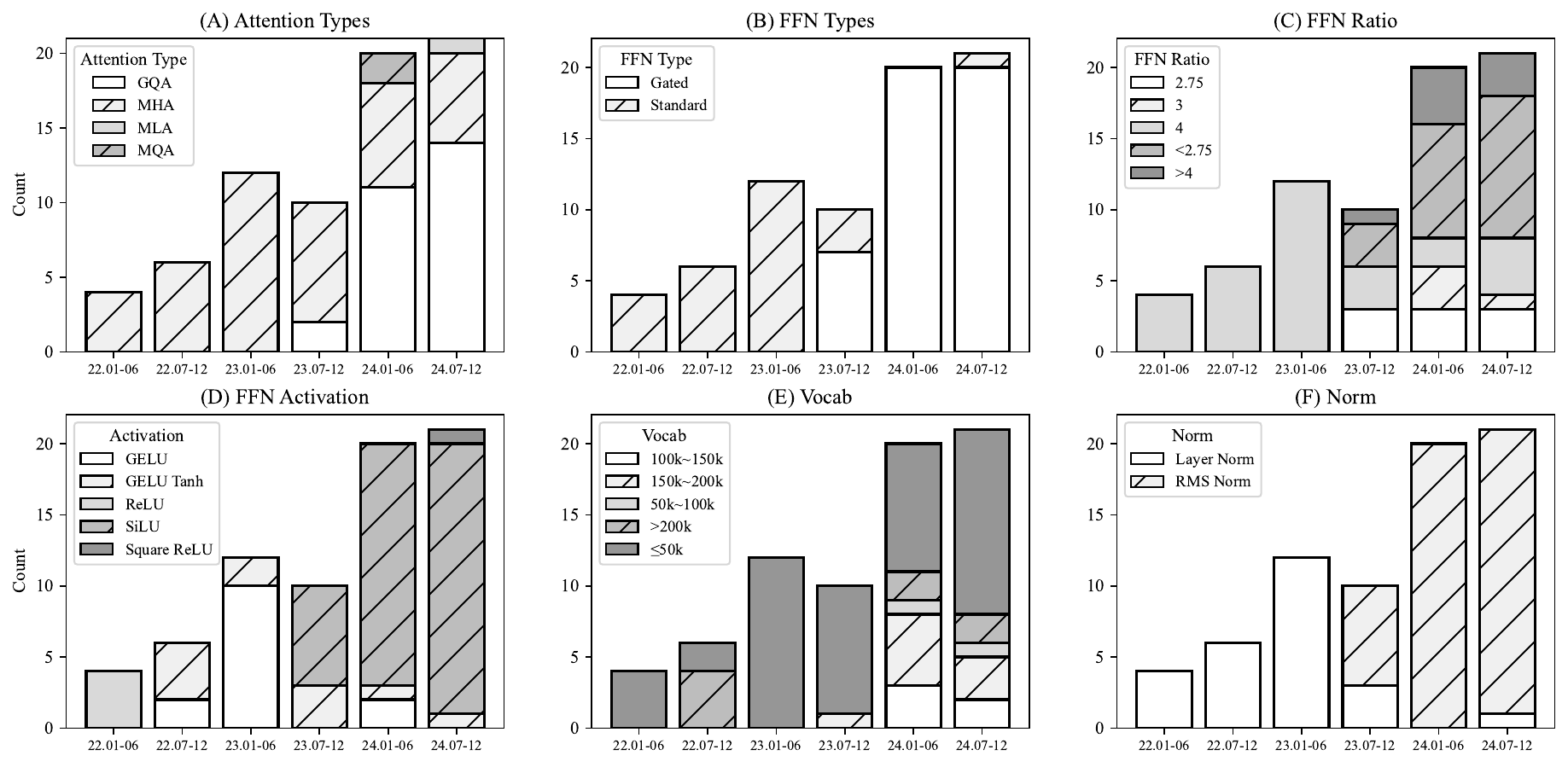}
        \subcaption{Architecture distribution. 
        }
    \end{minipage}
    \caption{
        The architecture analysis of the SLM, highlighting 6 configurations: attention type, FFN type, FFN ratio, FFN activation, vocabulary size, and normalization type. 
        (a) presents the overall structure of the SLM, and the categorizations with usage frequency of the 6 configurations;
        (b) analyzes the concrete selections of six configurations over time.
    }
    \label{fig:innovation-model-arch}
\end{figure*}

%% file: sec-innovation-dataset.tex
\subsection{Training Datasets}\label{subsec:dataset}

We investigate how the open-sourced pre-training datasets are used in training the SLMs.
Overall, we find 12 such datasets being used:
\begin{itemize}
\item The Pile~\cite{gao2020pile} (825B tokens): a combination of smaller corpora in various domains.
\item FineWeb-Edu~\cite{penedo2024fineweb} (1.3T tokens): a collection of educational text filtered from FineWeb.
\item StarCoder~\cite{li2023starcoder} (35B tokens): Python tokens.
\item Cosmopedia~\cite{benallal2024cosmopedia} (25B tokens): a dataset of synthetic textbooks, blogposts, stories, posts and WikiHow articles generated by Mixtral-8x7B-Instruct-v0.1.
\item RefinedWeb~\cite{penedo2023refinedweb} (5T tokens): despite extensive filtering, high-quality data extracted from the web remains plentiful, obtained from CommonCrawl.
\item RedPajama~\cite{together2023redpajama} (1.2T tokens):  includes over 100B text documents coming from 84 CommonCrawl snapshots and processed using the CCNet pipeline.
\item Dolma~\cite{dolma}: a English corpora, which is deduplicated inner corpus and across corpus using MinHash algorithms.
\item WuDaoCorpora~\cite{yuan2021wudaocorpora} (4T tokens): a super large-scale Chinese corpora, containing about 3T training data and 1.08T Chinese characters.
\item RoBERTa~\cite{liu2019roberta} CCNewsV2: containing an updated version of the English portion of the CommonCrawl News dataset.
\item PushShift ().io Reddit~\cite{baumgartner2020pushshift}: a social media data collection, analysis, and archiving platform that since 2015 has collected Reddit data and made it available to researchers.
\item DCLM-baseline~\cite{li2024datacomplm} (1.35T tokens): a standardized corpus extracted from Common Crawl, effective pretraining recipes based on the OpenLM framework, and a broad suite of 53 downstream evaluations.
\item CulturaX~\cite{nguyen2023culturax} (6.3T tokens): a substantial multilingual dataset in 167 languages. 
\end{itemize}




\textbf{The usage preference of pre-training datasets.}
\input{figs-innovation/fig-innovation-train-datasets.tex}
We then conducted statistics on the usage frequency of the datasets for training SLM from 2022 to 2024.
The results are illustrated in Figure \ref{fig:innovation-train-datasets}.
It shows that The Pile is the most widely used pre-training dataset especially in 2022 and 2023; yet more recently, more such datasets are proposed and the choice becomes diversified.
In fact, The Pile has been abandoned in pre-training SLMs recently, and datasets such as "RefinedWeb" and "RedPajama" have gradually been widely used.
It shows the active research and engineering efforts in constructing pre-training datasts with better quality.

\input{figs-innovation/tab-innovation-dataset-metric.tex}

\textbf{Comparing the quality of pre-training datasets.}
We also studied the open-sourced pre-training datasets quality based on the performance of SLMs trained on them.
We classified the SLMs in the past three years into groups of less than 0.5B, 1B, 2B, and 3B according to the number of parameters, and sorted them based on the average accuracy (the average of the accuracy of the two types of tasks, Commonsense reasoning/understanding and Problem solving, as will be shown in later section) of the four groups of models to explore how to select datasets to improve the average accuracy. The results are shown in Table \ref{tab:innovation-dataset-metric}. 
We notice that two recently released datasets, DCLM and FineWeb-Edu, show superior performance compared to others.
One common feature of the two datasets is the adoption of model-based data filtering.
In addition, coding data is often included in the SLM pretraining datasets such as StarCoder, even though coding ability is not a focus of SLMs deployed on devices.
This is likely attributed to the common belief that coding data can help improve the model reasoning ability~\cite{zhang2024unveiling}.

\textbf{The number of training tokens vs. the size of model parameters.}
\input{figs-innovation/fig-innovation-train-datasets-distribution.tex}
The number of parameters in SLM models and the amount of data used for training (the number of tokens) are closely related, with the Chinchilla law~\cite{hoffmann2022training} suggesting that the optimal ratio between the number of model parameters and training tokens should be around 20 (e.g., a 1B model with 20B tokens). We have statistically analyzed the number of training tokens used by SLMs under 4B parameters from 2022 to 2024, as shown in Figure \ref{fig:innovation-train-datasets-distribution}(a). Generally, the larger the model, the greater the number of tokens used for training, and more recent models tend to have more training tokens. A notable observation is that SLMs are trained on much large number of tokens (typically over 1.5T) than what is suggested by the Chinchilla law, regardless of their parameter sizes. In some cases, smaller SLMs are trained on even more data than the larger SLMs (e.g., Qwen2-0.5B with 12T tokens compared to Qwen2-1.5B with 7T tokens). This indicates that these SLMs are significantly "over-trained". The rationale behind this approach is to deploy powerful SLMs on resource-constrained devices by using more training-time FLOPs.
Though, SLMs are known to have performance saturation issue with over-training~\cite{godey2024small}.

\textbf{The amount of training tokens vs. model accuracy.}
Figure \ref{fig:innovation-train-datasets-distribution}(b) shows the relationship between the number of training tokens and the accuracy of the model.
In general, there is a positive correlation between the two metrics, especially for those with less than 700B tokens.
However, the correlation is weak, since the data quality often outweighs the impacts of more training tokens, especially when the training tokens exceed 1T.


\begin{remark}
\textbf{Insights}: We make two key observations in SLM training datasets.
\begin{itemize}[leftmargin=*]
    \item Data quality is crucial to SLM capability, which receives increasing attentions in recent SLM research.
    The importance of data quality to the final SLM capability typically outweighs the data quantity and model architecture configurations.
    A notable trend of dataset research is using model-based filtering, which result in two state-of-the-art open-sourced pre-training datasets: FineWeb-Edu (1.3T/5.4T)~\cite{penedo2024fineweb} and DCLM-baseline (4T)~\cite{li2024datacomplm}. SLMs trained on these two datasets have achieved competitive performance to those trained on closed datasets, which have significantly advanced the SLM research reproducibility.
\end{itemize}
\end{remark}
\begin{remark}
\begin{itemize}[leftmargin=*]
\item Recent SLMs are trained over large amount of tokens (typically $>$1.5T), disregarding their parameter sizes.
    In some cases, smaller SLMs are trained over even more data (e.g., Qwen2-0.5B at 12T tokens but Qwen2-1.5B at 7T tokens).
    It also means those SLMs are significantly ``over-trained'', as compared to the Chinchilla law~\cite{hoffmann2022training} that estimates the parameter-token ratio to be around only 20 (e.g., 1B model with 20B tokens).
    The incentive of such ``over-training'' action is to deploy powerful SLMs on resource-constrained devices through investing more training-time compute resources.
\end{itemize}
\end{remark}

%% file: figs-innovation/fig-innovation-train-datasets.tex
\begin{figure}[t]
    \centering
    \includegraphics[width=\textwidth]{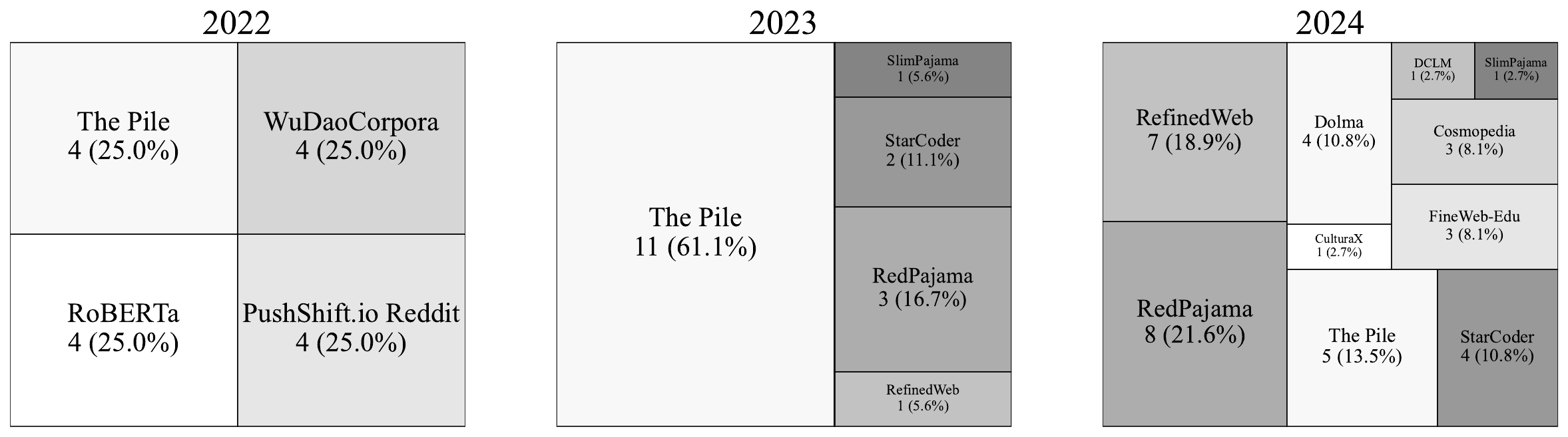}
    \vspace{-20pt}	
    \caption{
        The usage frequency of each open-source pre-training dataset from 2022 to 2024  
    }
    \label{fig:innovation-train-datasets}
\end{figure}

%% file: figs-innovation/tab-innovation-dataset-metric.tex
\begin{table}[t]
    \footnotesize
    \centering
    \begin{tabularx}{\textwidth}{llrrXr}
        \toprule
        Subcaption & Model & Date & Tokens(B) & Datasets & Acc(Avg) $\downarrow$ \\
        \midrule
        \multirow{11}{*}{\textless1B} 
        & SmolLM-360M & 24.07 & 600 & FineWeb-Edu\textsuperscript{b},StarCoder,Cosmopedia\textsuperscript{a} & 0.448 \\
        & OpenELM-450M & 24.04 & 1500 & RefinedWeb, The Pile, RedPajama, Dolma & 0.417 \\
        & SmolLM-135M & 24.07 & 600 & FineWeb-Edu\textsuperscript{b},StarCoder,Cosmopedia\textsuperscript{a} & 0.416 \\
        & MobiLlama-0.5B & 24.02 & 1259 & RedPajama, RefinedWeb & 0.405 \\
        & OpenELM-270M & 24.04 & 1500 & RefinedWeb, The Pile, RedPajama, Dolma & 0.393 \\
        & Pythia-410M & 23.03 & 300 & The Pile & 0.388 \\
        & BLOOMZ-560M & 22.11 & 350 & WuDaoCorpora & 0.366 \\
        & BLOOM-560M & 22.11 & 350 & WuDaoCorpora & 0.363 \\
        & OPT-125M & 22.05 & 180 & RoBERTa, The Pile, PushShift.io Reddit & 0.361 \\
        & Cerebras-GPT-590M & 23.03 & 12 & The Pile & 0.358 \\
        & OPT-125M & 22.05 & 180 & RoBERTa, The Pile, PushShift.io Reddit & 0.349 \\
        & Pythia-160M & 23.03 & 300 & The Pile & 0.347 \\
        & Cerebras-GPT-111M & 23.03 & 2 & The Pile & 0.330 \\
        \midrule
        \multirow{8}{*}{1B--1.4B} 
        & DCLM-1B & 24.08 & 4300 & DCLM-baseline\textsuperscript{b} & 0.577 \\
        & OpenELM-1.1B & 24.04 & 1500 & RefinedWeb, The Pile, RedPajama, Dolma & 0.463 \\
        & TinyLlama-1.1B & 23.12 & 3000 & SlimPajama, StarCoder & 0.436 \\
        & MobiLlama-1B & 24.02 & 1259 & RedPajama, RefinedWeb & 0.434 \\
        & MobileLLaMA-1.4B & 23.12 & 1300 & RedPajama & 0.428 \\
        & Pythia-1.4B & 23.03 & 300 & The Pile & 0.423 \\
        & OPT-1.3B & 22.05 & 180 & RoBERTa, The Pile, PushShift.io Reddit & 0.413 \\
        & Pythia-1B & 23.03 & 300 & The Pile & 0.406 \\
        & Bloom-1B1 & 22.11 & 350 & WuDaoCorpora & 0.394 \\
        & Bloomz-1B1 & 22.11 & 350 & WuDaoCorpora & 0.384 \\
        & Cerebras-GPT-1.3B & 23.03 & 26 & The Pile & 0.383 \\
        \midrule
        \multirow{2}{*}{1.5B--2B}
        & StableLM-2-zephyr-1.6B & 24.01 & 2000 & RefinedWeb, RedPajama, The Pile, StarCoder, CulturaX & 0.556 \\
        & SmolLM-1.7B & 24.07 & 1000 & FineWeb-Edu\textsuperscript{b},StarCoder,Cosmopedia\textsuperscript{a} & 0.503 \\
        \midrule
        \multirow{3}{*}{2.5B--3B} 
        & StableLM-zephyr-3B & 23.11 & 400 & RefinedWeb, RedPajama, The Pile, StarCoder& 0.582 \\
        & Pythia-2.8B & 23.03 & 300 & The Pile & 0.448 \\
        & OPT-2.7B & 22.05 & 180 & RoBERTa, The Pile, PushShift.io Reddit & 0.439 \\
        & Cerebras-GPT-2.7B & 23.03 & 53 & The Pile & 0.405 \\
        \bottomrule
    \end{tabularx}
    \caption{Classify according to the model parameter quantity and sort in descending order according to average normalized accuracy.
    Acc(Avg) is the average of the accuracies of the two types of tasks, Commonsense reasoning/understanding and Problem solving.
    \textbf{a} indicates that this dataset is generated by LLM.
    \textbf{b} indicates that this dataset has been filtered by LLM.
    }
    \vspace{-20pt}
    \label{tab:innovation-dataset-metric}
\end{table}

%% file: figs-innovation/fig-innovation-train-datasets-distribution.tex
\begin{figure}[t]
    \centering
    \begin{minipage}[b]{0.48\textwidth}
        \vspace{0pt} 
        \includegraphics[width=\textwidth]{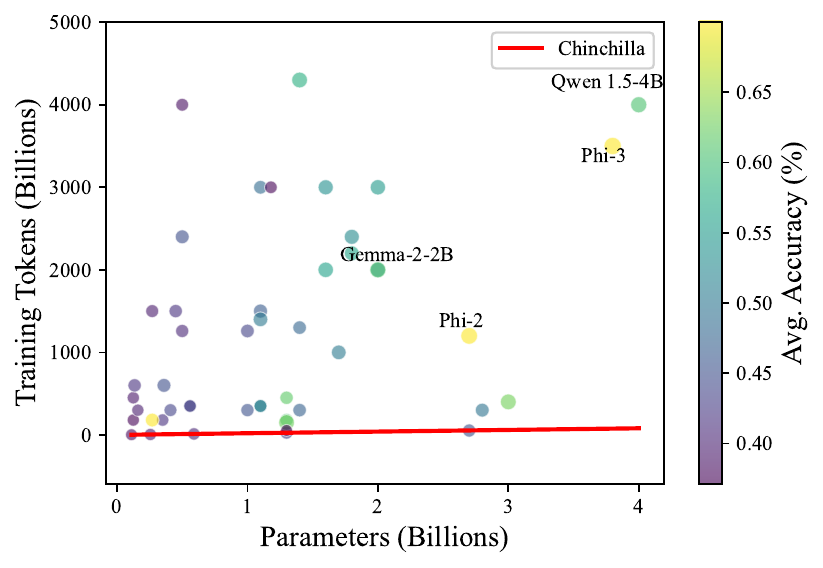}
        \subcaption{The relationship between Training Tokens and Parameters. }
    \end{minipage}
    \hspace{0.0\textwidth} 
    \begin{minipage}[b]{0.48\textwidth}
        \includegraphics[width=\textwidth]{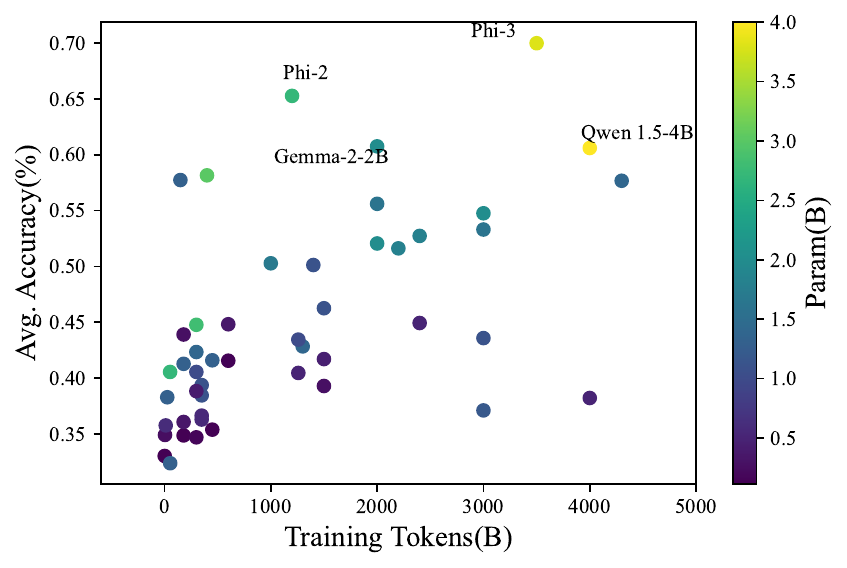}
        \subcaption{The influence of Training Tokens on Accuracy}
    \end{minipage}
    \vspace{-5pt}	
    \caption{
        The relationship between the number of training tokens, the number of model parameters, and the model accuracy. 
        Here, the ``accuracy'' is averaged across all benchmarks in Table~\ref{tab:innovation-dataset-metric}.
        (a) The relationship between the number of training tokens and model parameters size. According to scaling law(Chinchilla), that SLMs are often over-trained for better performance at deployment stage.
        (b) The influence of the number of training tokens on the model accuracy.
    }
    \vspace{-10pt}	
    \label{fig:innovation-train-datasets-distribution}
\end{figure}

%% file: sec-innovation-algorithm.tex
\subsection{Training Algorithms}


There have been a few novel training methods to improve the model capability.

\textbf{Maximal Update Parameterization($\mu$P)} controls initialization, layer-wise learning rates, and activation magnitudes to ensure analytically stable training independent of a model’s layer widths. In addition to improving training stability, $\mu$P also improves the transferability of training hyperparameters from smaller to larger scale models, which permits directly using the same settings for some optimizer hyperparameters, most notably the learning rate. For example, Cerebras-GPT trains models with Maximal Update Parameterization.

\textbf{Knowledge Distillation} is a crucial concept in the realm of Large Language Models (LLM). 
It involves extracting valuable knowledge from a large and complex teacher model and transferring it to a smaller and more efficient student model.
The essence of this technique is to have the student model learn to approximate the behavior and predictions of the teacher. This is achieved by minimizing the difference between their outputs.
According to our statistics, LaMini-GPT and Gemma-2 adopt Knowledge Distillation.

\textbf{Two Stage Pre-training Strategy} is a training strategy that involves training a model in two distinct phases.
During the pretraining phase, MiniCPM only uses large-scale coarse-quality pre-training data, which is abundant and can support continuous training when provided with more computational resources. During the annealing phase, we use diverse and high-quality knowledge and ability-oriented SFT data, mixed into the pre-training data.
MninCPM adopts Two Stage Pre-training Strategy. 

%% file: sec-capability.tex
\section{SLM Capabilities}

\subsection{Evaluation Datasets and Metrics}

We used 12 datasets across three domains to evaluate the SLM performance.

\begin{itemize}
    \item \textbf{Commonsense Reasoning Datasets}:
    \begin{itemize}
    \item \textbf{HellaSwag}~\cite{zellers2019hellaswag}: Tests narrative understanding through plausible sentence completion.
    \item \textbf{TruthfulQA}~\cite{lin2022truthfulqa}: Assesses the model's ability to avoid providing false information.
    \item \textbf{Winogrande}~\cite{sakaguchi2020winogrande}: Evaluates pronoun ambiguity resolution using commonsense reasoning.
    \item \textbf{CommonsenseQA}~\cite{talmor2019commonsenseqa}: Presents multiple-choice questions requiring everyday knowledge.
    \item \textbf{PIQA}~\cite{bisk2020piqa}: Focuses on physical commonsense reasoning and object interactions.
    \item \textbf{OpenBookQA}~\cite{mihaylov2018can}: Combines scientific knowledge with commonsense for open-book science questions.
    \item \textbf{BoolQ}~\cite{clark2019boolq}: Tests commonsense and factual reasoning with yes/no questions.
    \end{itemize}
    \item \textbf{Problem-Solving Datasets}:
    \begin{itemize}
    \item \textbf{ARC Easy}~\cite{clark2018think}: Contains simple science questions testing general knowledge and reasoning.
    \item \textbf{ARC Challenge}~\cite{clark2018think}: Presents complex science exam questions requiring knowledge integration.
    \item \textbf{MMLU}~\cite{hendrycks2021measuring}: Evaluates problem-solving across diverse academic disciplines.
    \end{itemize}
    \item \textbf{Mathematics Datasets}:
    \begin{itemize}
    \item \textbf{GSM8K}~\cite{cobbe2021training}: Assesses grade-school-level mathematical reasoning skills.
    \item \textbf{Minerva Math}~\cite{lewkowycz2023minerva}: Evaluates advanced mathematical reasoning across various topics.
    \end{itemize}
    \end{itemize}

We use \textit{accuracy} as the primary evaluation metric.
Accuracy measures the proportion of correct predictions to total examples.
The default shown accuracy is instructed by 5 shots, as it is the most common setting in the released model.
For commonsense reasoning, problem-solving, and mathematics tasks, accuracy evaluates the model's ability to select correct options or provide accurate solutions.

\subsection{Overall Capabilities}
As shown in Figure~\ref{fig:acc-evolution}, we conducted experiments on selected SLMs across three tasks—commonsense reasoning, problem-solving, and mathematics—to analyze their progress. 
The results show substantial performance improvements across all tasks between 2022 and 2024. 
Specifically, model performance improved by 10.4\%, 13.5\%, and 13.5\% for the three tasks, respectively. 
In comparison, the state-of-the-art open-source LLaMA model exhibited an average improvement of only 7.5\% over the same period. 
Notably, the \texttt{Phi} family, trained on closed-source datasets, outperforms all other models, achieving 67.6\% in commonsense reasoning and 72.4\% in problem-solving—levels comparable to the latest LLaMA 3.1 with 7 billion parameters. 
For example, in the mathematics task, \texttt{Phi-3-mini} demonstrates a substantial 14.5\% lead over LLaMA 3.1. 
These results suggest that SLMs are rapidly closing the gap with LLMs in general reasoning tasks, although some differences remain, particularly in mathematics. 
Moreover, while larger parameter counts generally correlate with better performance, there are notable exceptions, such as Qwen 2, which outperforms many SLMs with 3 billion parameters despite having only 1.5 billion.

\input{fig-acc-evolution.tex}

Although pioneering SLMs are trained on closed-source datasets, the gap between open-source and closed-source trained models in commonsense tasks is narrowing. 
For example, SmolLM and DCLM-1B perform exceptionally well in commonsense reasoning (achieving 64.2\% and 63.8\%, respectively), thanks to high-quality datasets such as DCLM and FineWeb-Edu. 
However, the gap remains significant in tasks requiring complex reasoning or logic, particularly in mathematics, likely due to the lack of high-quality logic datasets. 

\begin{remark}
    \textbf{Insights}: 
    We draw four key insights from the development of SLMs:
    \begin{itemize}
        \item From 2022 to 2024, SLMs exhibited significant performance improvements across various language tasks, outpacing the improvements of the LLaMA-7B series (1/2/3/3.1 versions).
        This paints a promising picture for SLMs' potential to solve a range of downstream tasks on devices.
        \end{itemize}
\end{remark}
\begin{remark}
    \begin{itemize}
        \item The \texttt{Phi} family consistently achieves state-of-the-art performance across most tasks. In particular, \texttt{Phi-3-mini} achieves the highest accuracy as of September 2024, rivaling LLaMA 3.1 8B.
        While much of its superior performance may be due to careful data engineering by the Microsoft team, part of it may also be attributed to instructive tuning and potential overfitting to specific datasets~\cite{zhang2024careful}.
        \item Although larger parameter counts generally lead to better performance, exceptions such as Qwen 2~1.5B demonstrate that smaller models can still excel in specific tasks.
        \item SLMs trained on open-source datasets are closing the gap with their closed-source counterparts in commonsense tasks. 
        However, the gap remains significant in tasks requiring complex reasoning or logic. 
        This underscores the need for improved datasets focused on mathematical reasoning to address this disparity.
    \end{itemize}
\end{remark}


\subsection{In-context Learning Capabilities}
\input{fig-acc-in-context.tex}

We conduct in-context learning experiments using various models and their 2B-parameter variants (or the closest available ones) across 8 tasks, including commonsense reasoning and problem-solving tasks. 
Generally, SLMs benefit significantly from in-context learning across all tasks. 
Exceptions include the HellaSwag and PIQA datasets, where all models perform similarly regardless of the number of in-context learning shots. 
These datasets are simpler and do not benefit as much from in-context learning as more complex datasets, such as ARC Challenge. 
On average, in-context learning with 5 shots improves the performance of zero-shot SLMs by 2.1\% across all tasks. 
The only notable exception is LaMini, which shows a decrease of over 2\% in performance. 
We hypothesize that this model may be overfitting the training data, and additional context shots introduce noise. 
Among the models, Gemma 2 exhibits the most significant improvement, with a 4.8\% increase in accuracy. 
Interestingly, we observe that as model size increases, the in-context learning capability of SLMs is enhanced.



\begin{remark}
    \textbf{Insights}: 
    We draw two key insights from the in-context learning capacity of SLMs:
    \begin{itemize}
        \item Generally, most SLMs encompass certain levels of in-context learning ability.
        However, such ability varies across different tasks: almost all SLMs benefit significantly from in-context learning in arc\_challenge task while certain tasks show mere benefit from in-context learning across all the models, such as hellaswag and piqa.
        \item Larger models tend to exhibit stronger in-context learning capabilities compared to their smaller counterparts. Some small SLMs even show a decrease in performance with in-context learning.
    \end{itemize}
\end{remark}

\subsection{Long Context Capabilities}
\input{fig-acc-long-context}
We used Needle-In-A-Haystack provided by OpenCompass to explore long-context capabilities of SLMs. The tasks included Single-Needle Retrieval, Multi-Needle Retrieval, and Multi-Needle Reasoning. The scores in Figure 7 are the average of these three tasks. Different models showed large variations in performance.
Small models, such as Qwen1.5-0.5B and Qwen2-0.5B, performed less effectively. Qwen1.5-0.5B achieved an average accuracy of 22.13\%. Qwen2-0.5B performed slightly better, reaching 43.84\%. Qwen1.5-0.5B handled shorter contexts (9k-17k) relatively well. However, its accuracy dropped sharply with longer contexts. This was especially true for middle inserted text (Depth Percent from 20\% to 70\%).
Larger models performed much better. Llama3.2-3B had an average accuracy of 57.81\%. It worked well with shorter contexts but struggled with deeper insertions when contexts exceeded 25k tokens. 
Qwen2.5-3B achieved an average accuracy of 91.71\%. It maintained nearly perfect accuracy across all context lengths and insertion positions. This result highlights its strong ability to handle long contexts and complex scenarios.

\begin{remark}
    \textbf{Insights}: 
    We draw two key insights from the long context capacity of SLMs:
    \begin{itemize}
    \item Larger parameters are crucial for long-context capabilities. Small models, such as Qwen1.5-0.5B and Qwen2-0.5B, perform adequately on short-context tasks but experience a significant drop in recognition accuracy as the context length increases. In contrast, larger models, such as Qwen2.5-3B, excel with outstanding performance, maintaining near-perfect accuracy across all context lengths and insertion positions.
    \end{itemize}
    \begin{itemize}
    \item "Lost in the Middle" also occurs in small models. Compared to deep or front insertions, the accuracy of middle-position text (Depth Percent 20\%-70\%) is significantly lower.
    \end{itemize}
\end{remark}

%% file: fig-acc-evolution.tex
\begin{figure}[t]
		\centering
				\centering
				\hspace*{-5pt}
    \includegraphics[width=1.02\textwidth]{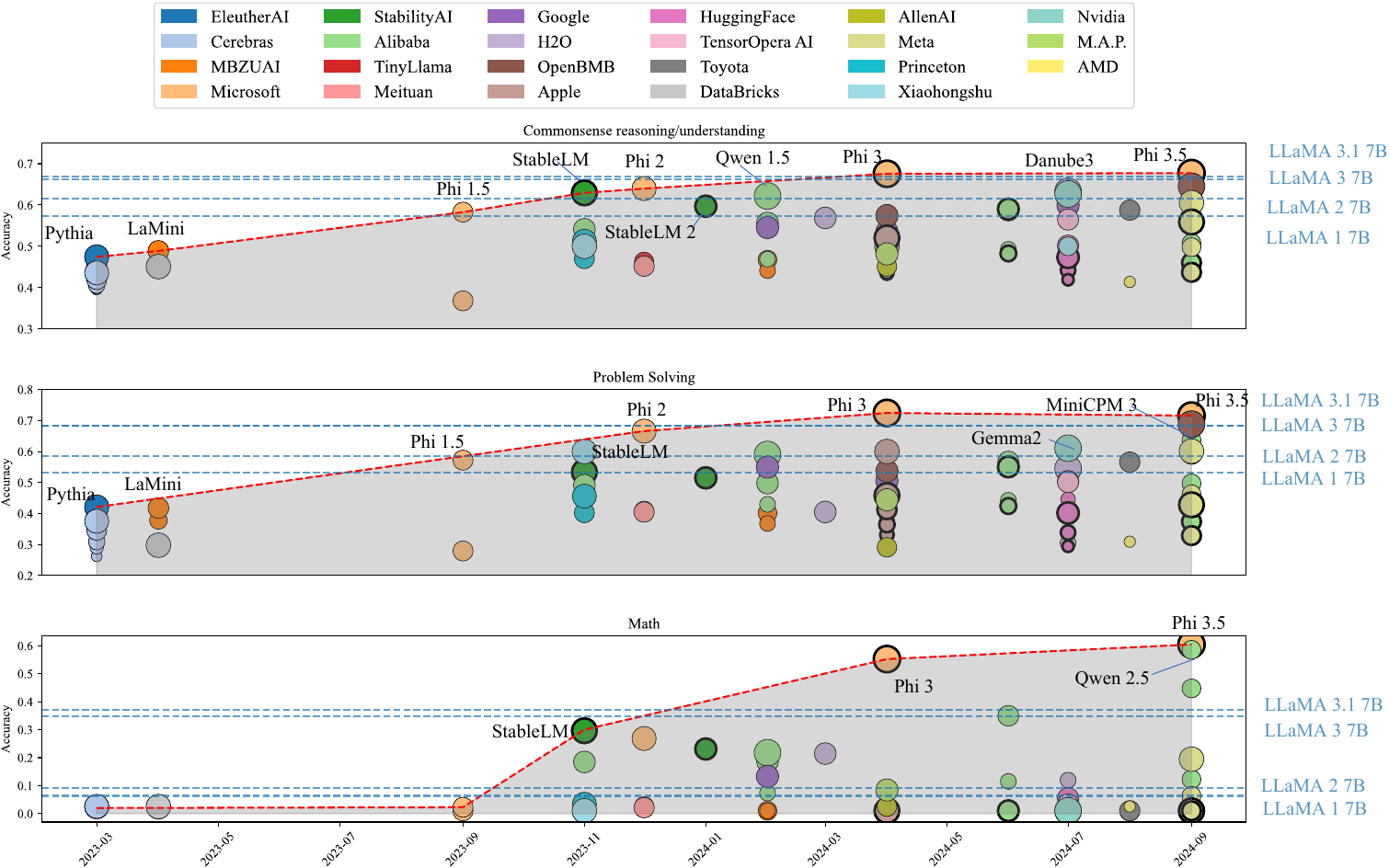}
				\label{fig:acc-evolution-1}
	\caption{
        SLM capabilities over time. 
		The size of the circle is proportional to the model size. 
		Red dashed lines show the state-of-the-art model at different time, indicating the trend that SLMs are getting better over time.
		LLaMA-7B series models are shown in horizontal blue dashed lines for comparison.
  		Note that Phi and StableLM series are instructed models, while others are base models.
		}
	\label{fig:acc-evolution}
\end{figure}

%% file: fig-acc-in-context.tex
\begin{figure}[h]
\centering
	\begin{minipage}
		{1.0\textwidth}
		\centering
		\includegraphics[width=0.98\textwidth]{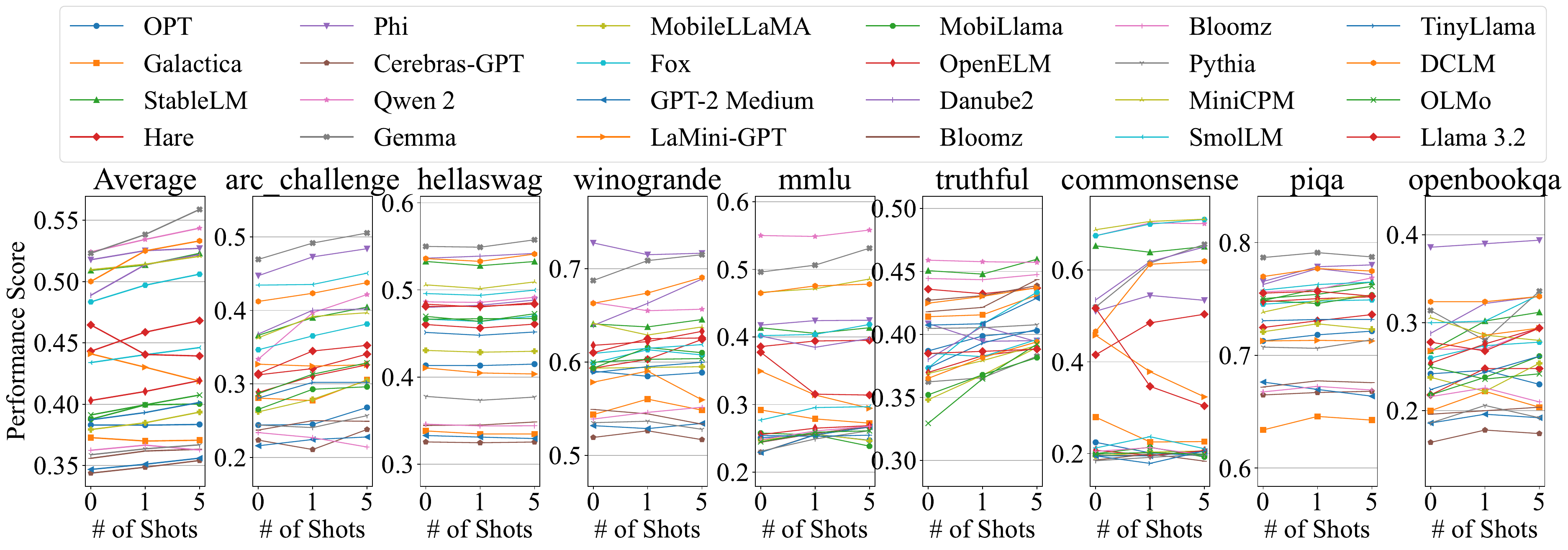}
		\label{fig:acc-incontext}
		\subcaption{
			SLM in-context capabilities across tasks.
			}
	\end{minipage}

	\begin{minipage}
		{1.0\textwidth}
		\centering
		\includegraphics[width=0.98\textwidth]{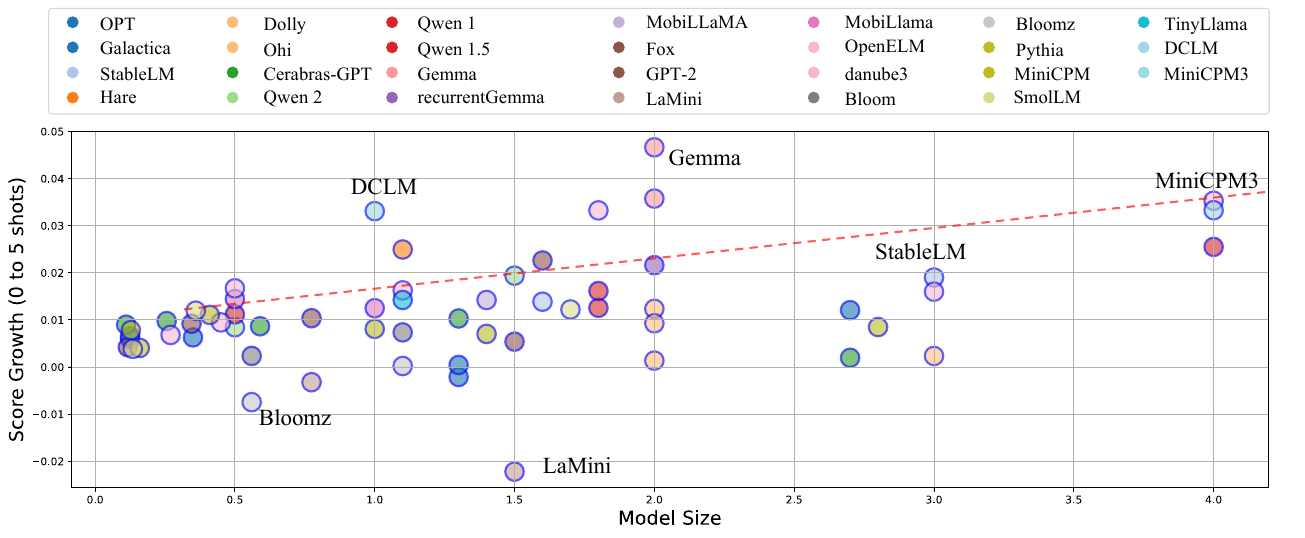}
		\subcaption{
			Average accuracy improvement after in-context learning across different SLM model size. 
			}
		\label{fig:acc-longcontext}
	\end{minipage}
	\caption{
			In-context learning performance with different tasks and models.
			Red line in (b) highlights the trend of the average score improvement with the increase of model size.
			}
\end{figure}

%% file: fig-acc-long-context.tex
\begin{figure}[h]
	\centering
    \centering
    \begin{minipage}[b]{0.23\textwidth}
        \vspace{0pt} 
        \includegraphics[width=\textwidth]{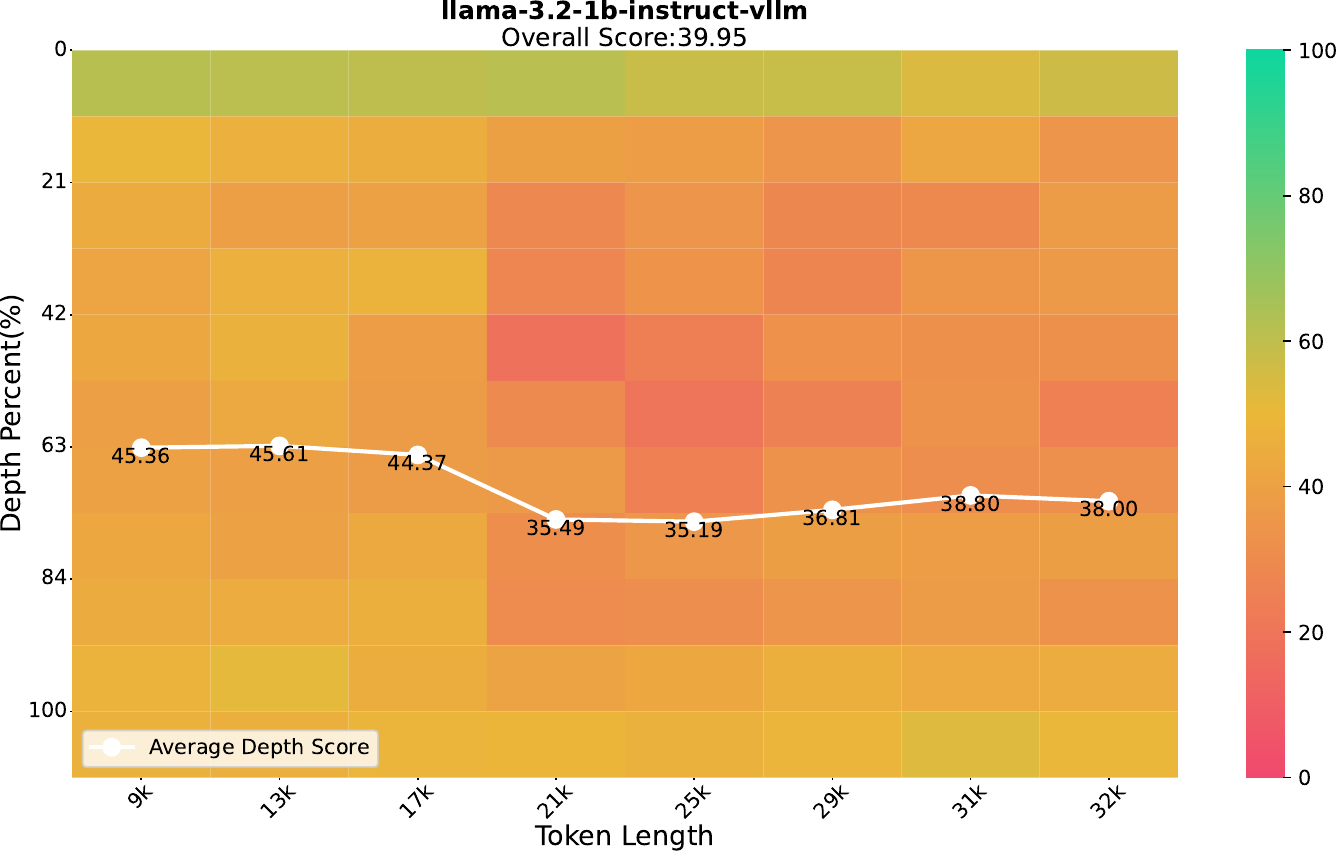}
        \subcaption{Llama3.2-1B}
    \end{minipage}
    \hspace{0.0\textwidth} 
    \begin{minipage}[b]{0.23\textwidth}
        \includegraphics[width=\textwidth]{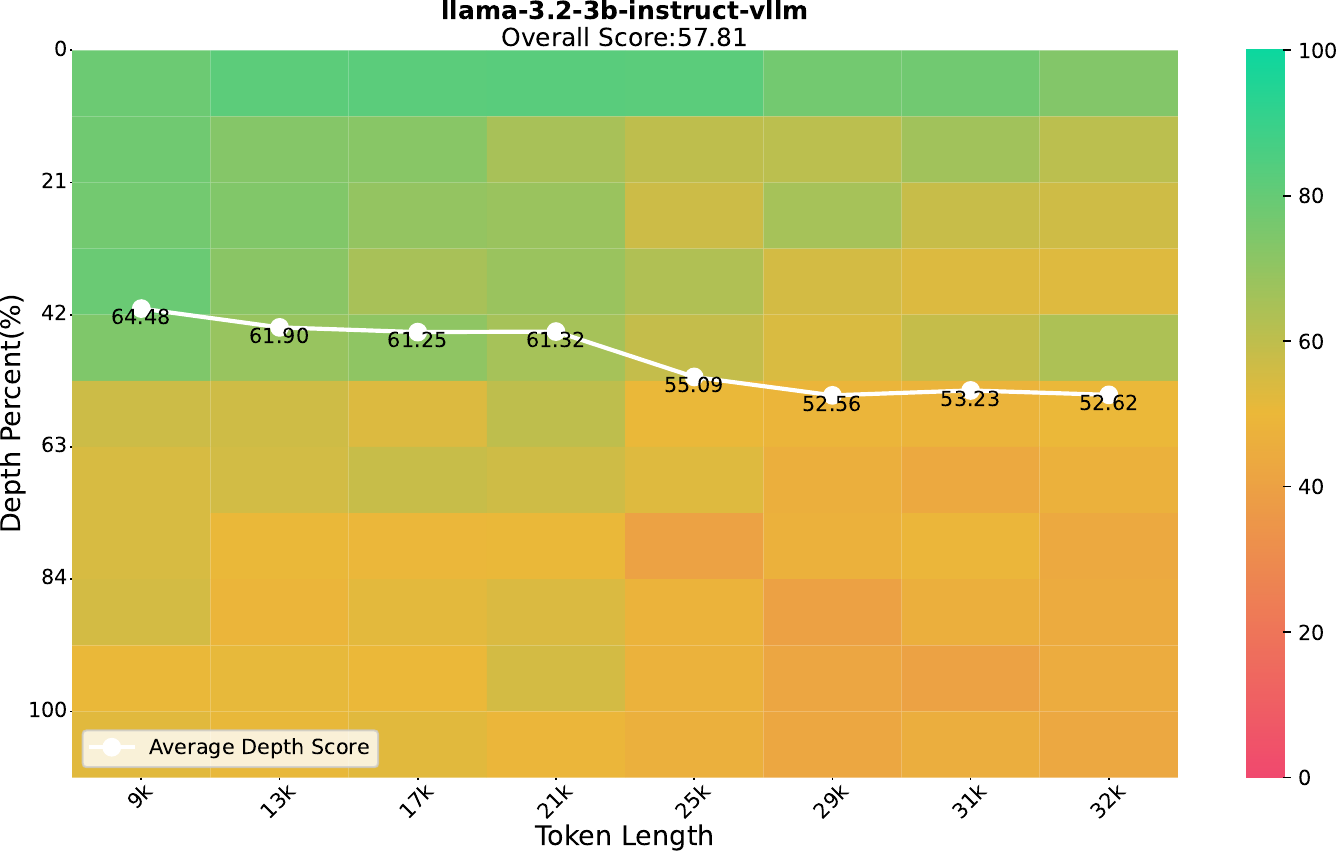}
        \subcaption{Llama3.2-3B}
    \end{minipage}
    \begin{minipage}[b]{0.23\textwidth}
        \vspace{0pt} 
        \includegraphics[width=\textwidth]{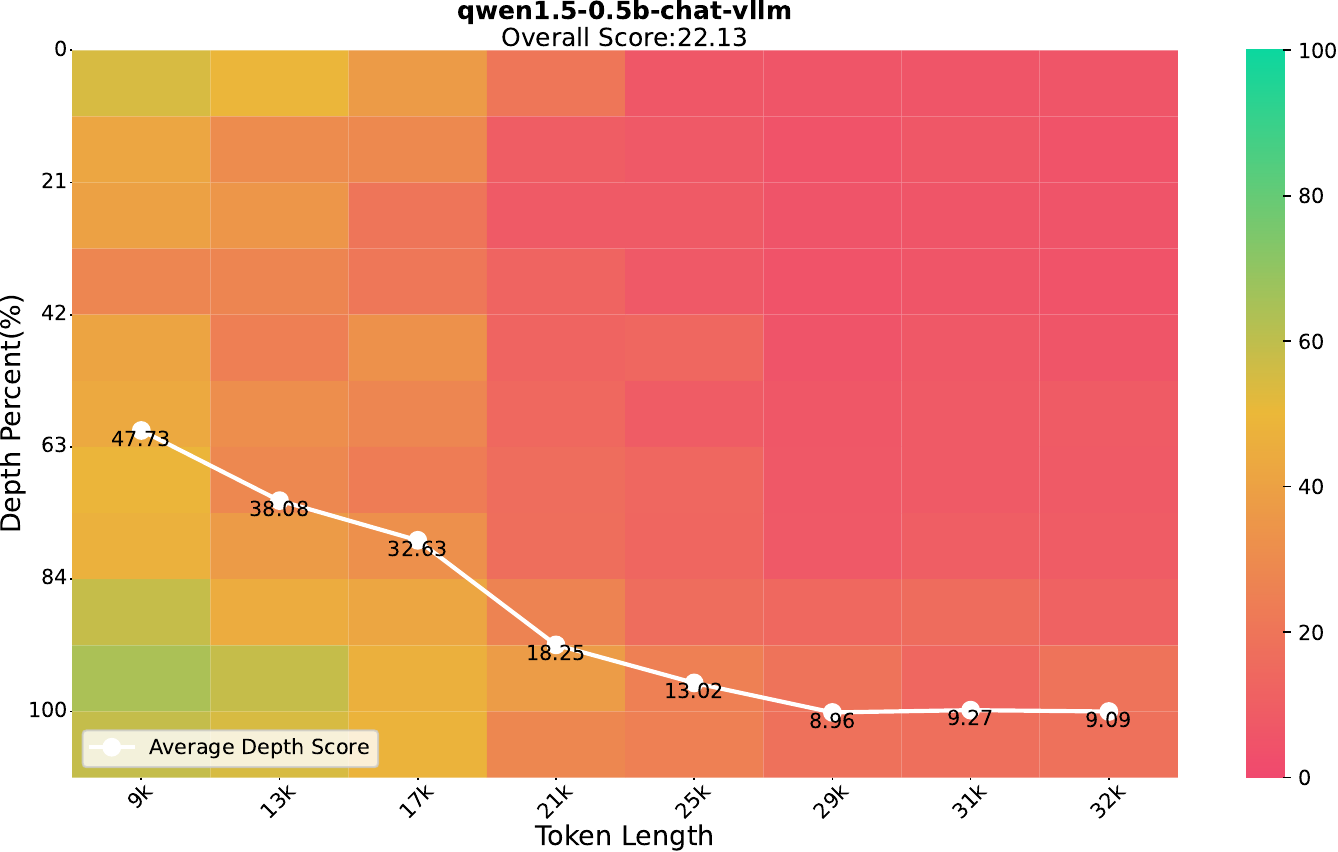}
        \subcaption{Qwen1.5-0.5B}
    \end{minipage}
    \hspace{0.0\textwidth} 
    \begin{minipage}[b]{0.23\textwidth}
        \includegraphics[width=\textwidth]{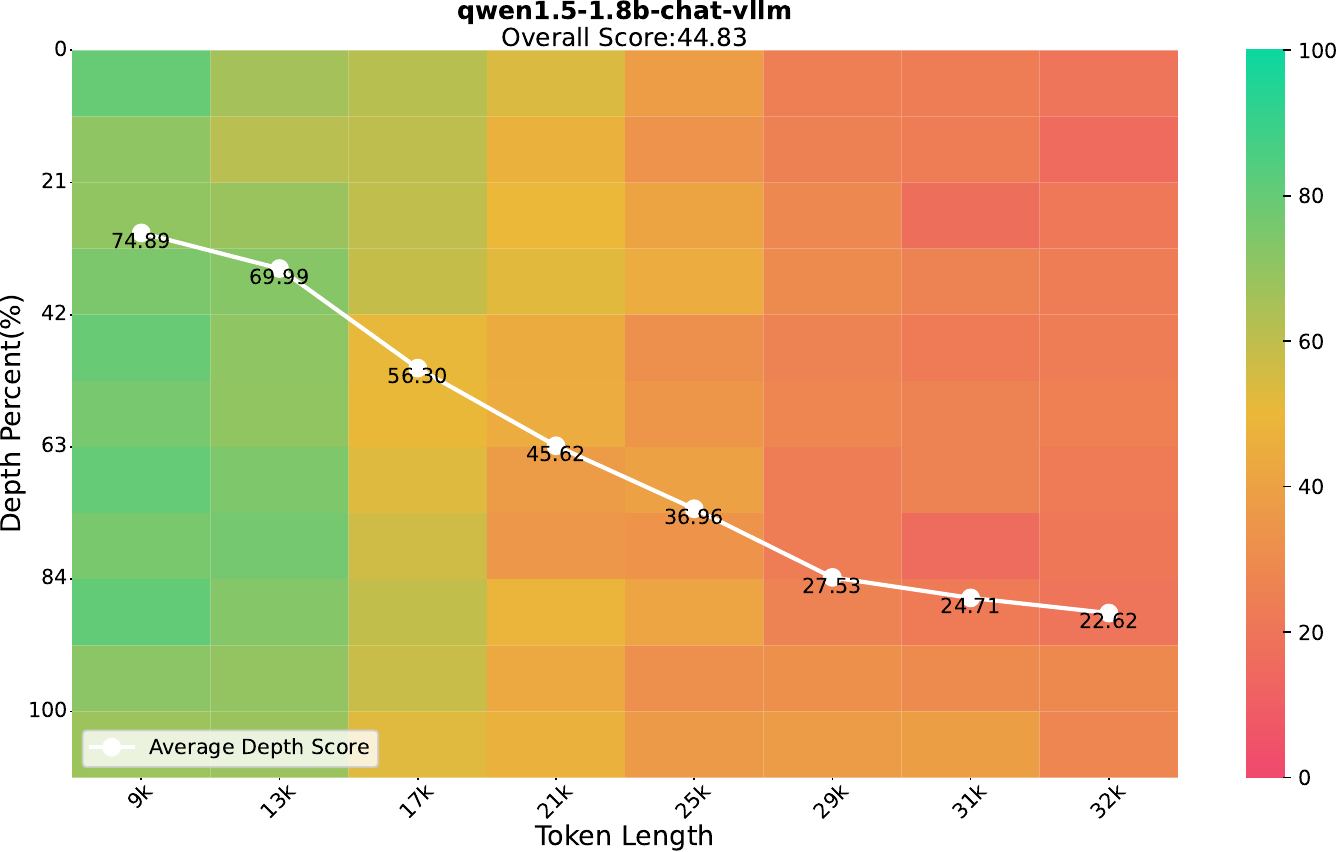}
        \subcaption{Qwen1.5-1.8B}
    \end{minipage}
    \begin{minipage}[b]{0.23\textwidth}
        \vspace{0pt} 
        \includegraphics[width=\textwidth]{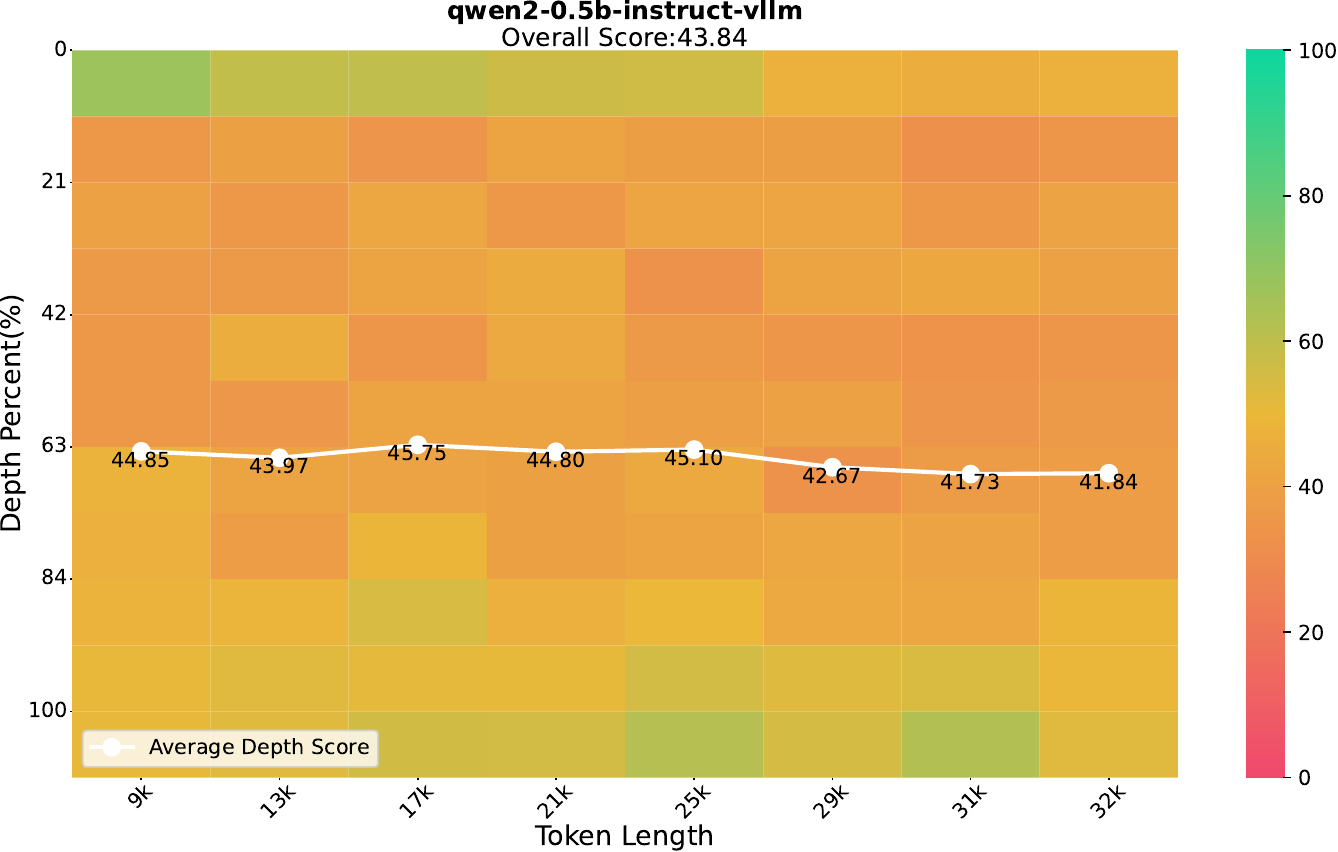}
        \subcaption{Qwen2-0.5B}
    \end{minipage}
    \hspace{0.0\textwidth} 
    \begin{minipage}[b]{0.23\textwidth}
        \includegraphics[width=\textwidth]{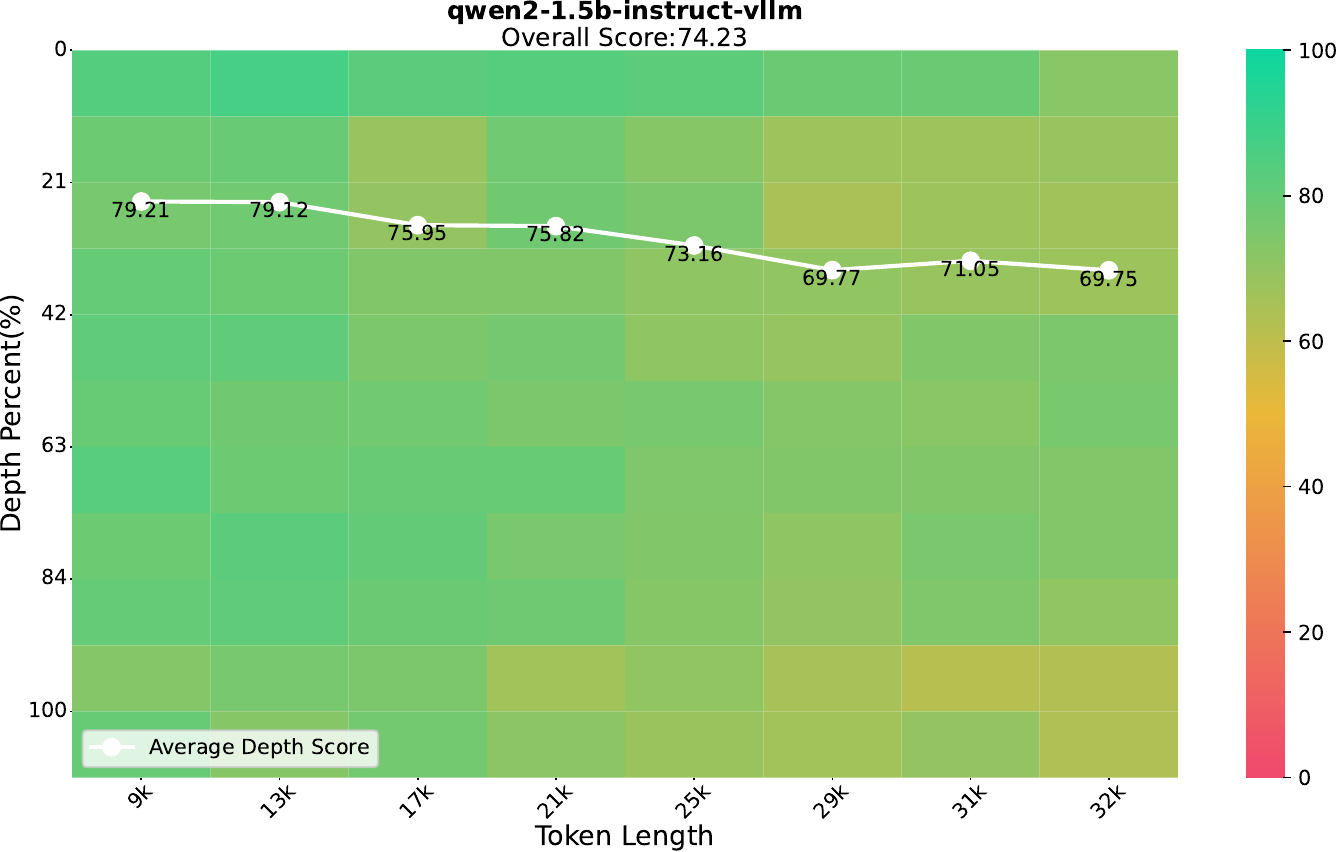}
        \subcaption{Qwen2-1.5B}
    \end{minipage}
    \begin{minipage}[b]{0.23\textwidth}
        \vspace{0pt} 
        \includegraphics[width=\textwidth]{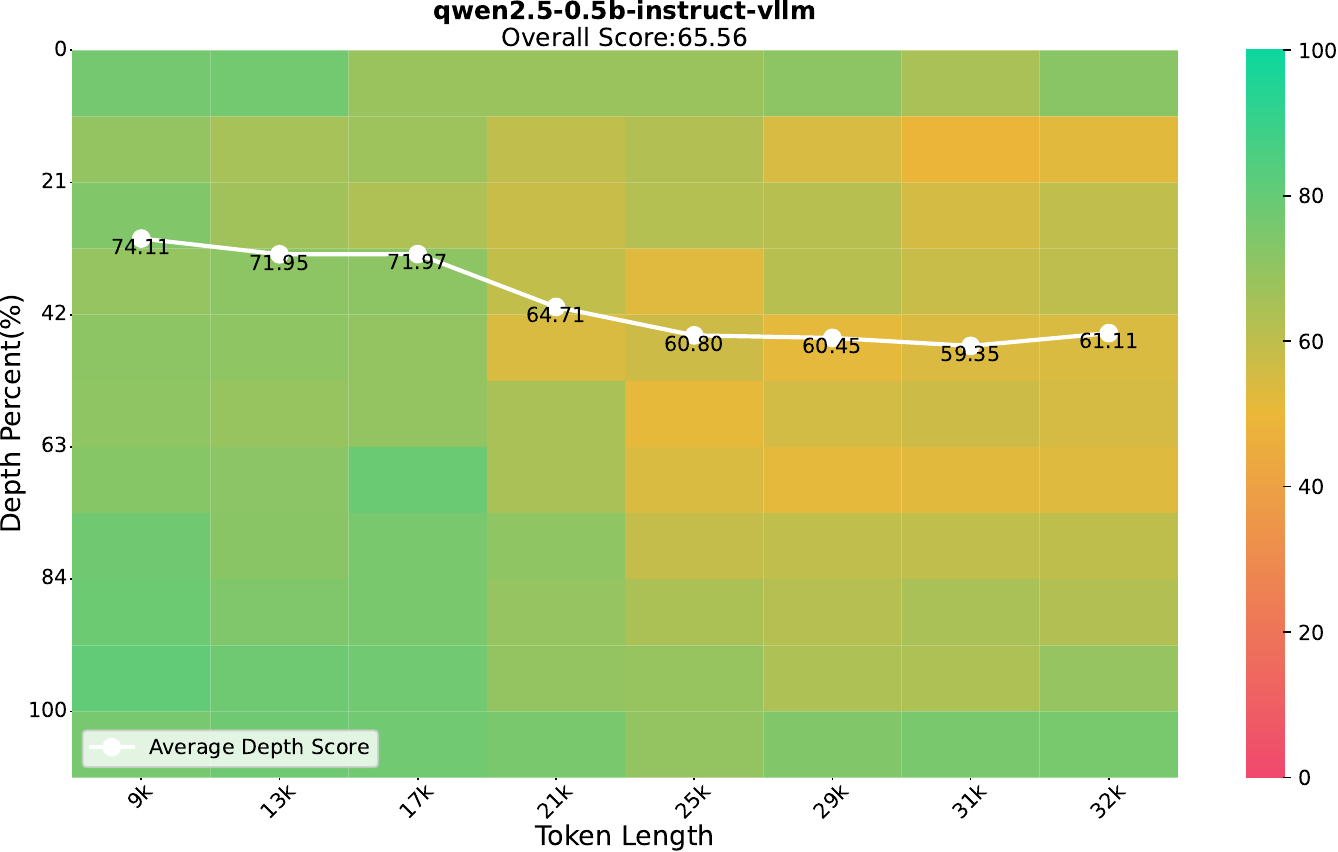}
        \subcaption{Qwen2.5-0.5B}
    \end{minipage}
    \hspace{0.0\textwidth} 
    \begin{minipage}[b]{0.23\textwidth}
        \includegraphics[width=\textwidth]{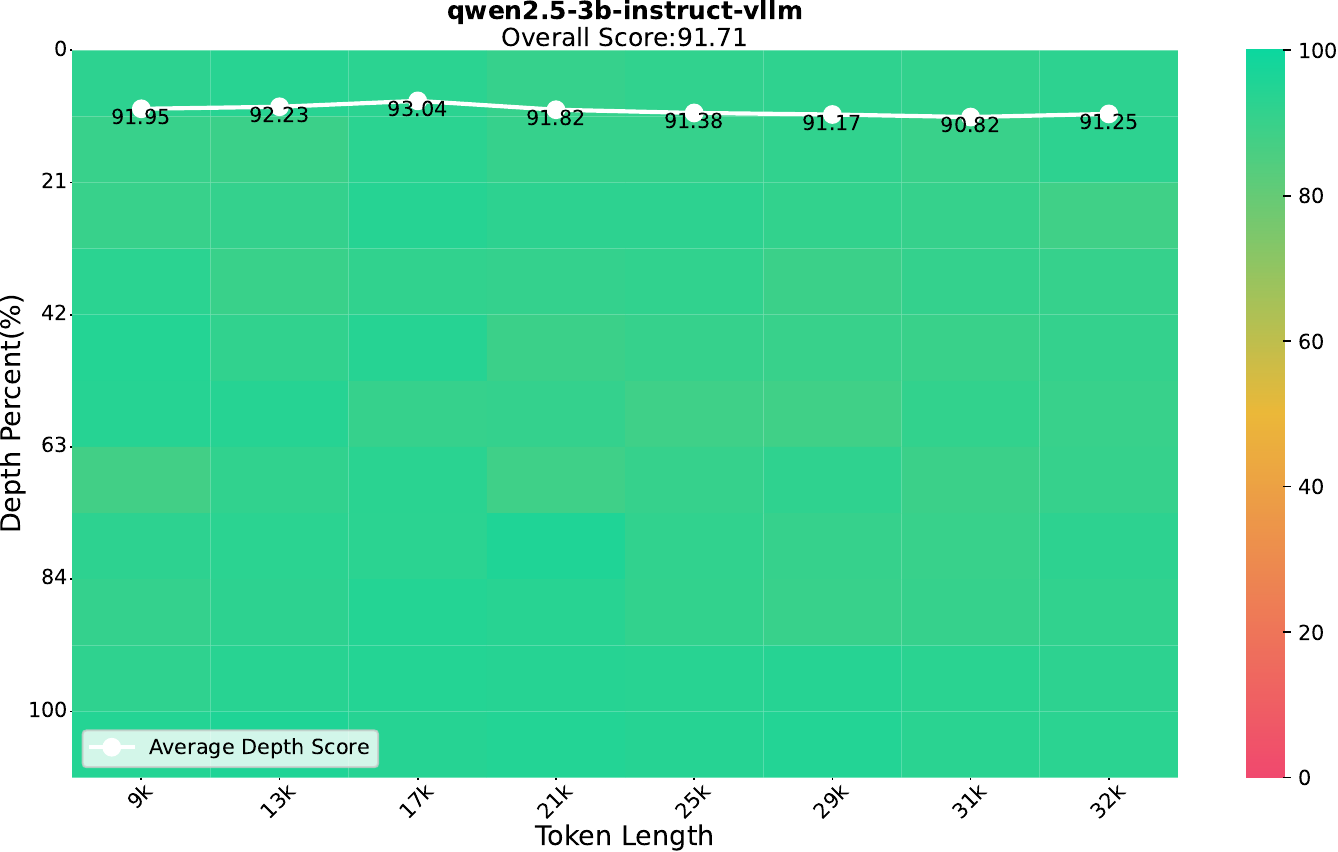}
        \subcaption{Qwen2.5-3B}
    \end{minipage}
    \vspace{0pt}	
    \caption{
        Needle In A Haystack
    }
	\label{fig:acc-longcontext}
\end{figure}

%% file: sec-cost.tex
\section{SLM Runtime Cost}\label{sec:cost}
\input{figs-cost/fig_model_size}



\textbf{Setup}
In this section, we first provide an overall analysis of the latency and memory used by models with different parameter sizes. Next, we examine the impact of quantization methods and hardware on model latency. Finally, we break down the latency and memory usage to identify the key factors affecting them in different parts of the model.

We evaluated 20 models on two types of edge devices: the Jetson Orin Module, commonly used in edge AI devices such as drones and small robots, and smartphones, which people rely on in their daily lives. The detailed specifications are shown in Table~\ref{tab:device_specs}. All experiments on the Jetson used its GPU, while those on the smartphones were performed using its CPU. To eliminate the impact of inference engine implementations, we carry out all experiments using \texttt{llama.cpp}, a widely recognized open-source inference engine. 
\begin{table}
    \footnotesize
\centering
\begin{tabular}{c|c|c}
\hline
\textbf{Device Name} & \textbf{Specifications}    & \textbf{Release Time} \\ \hline
Jetson Orin NX 16GB   & 1024-core NVIDIA Ampere
architecture GPU with 32
tensor cores, 16G DRAM           & Feb. 2023    \\ \hline
Pixel 7Pro  & GoogleTensor G2,12G RAM & Oct. 2022    \\ \hline
Xiaomi 12S  & Snapdragon 8Gen1+ ,12G RAM             & Jul. 2022    \\ \hline
MEIZU 18Pro & Snapdragon 888,8G RAM             & Mar. 2021     \\ \hline
\end{tabular}
\caption{Testing devices.}
\vspace{-6mm}
\label{tab:device_specs}
\end{table}

We primarily recorded metrics as model parameter, latency during the prefill and decode phases, and runtime memory usage. 
Due to variations in how each model officially reports its parameter counts, we relied on the parameter values obtained from llama.cpp. Inference is divided into two phases: prefilling and decoding.
During the prefill phase, the input prompt is processed to generate a KV Cache, where multiple tokens in the prompt can be computed in parallel. For the prefill phase, we focused on first token latency, which represents the time it takes to process all tokens in the prompt. 
The decode phase, also known as the autoregressive phase, generates one token at a time, incorporating it into the KV Cache. Simultaneously, this token is used in predicting the next token. For the decode phase, we measured latency per token.
We set a standard prompt length of 50 and a token generation length of 50 unless specified otherwise. Tests are conducted at 10-second intervals to mitigate potential thermal throttling issues.
To measure larger models, we applied 4-bit quantization to all models before conducting experiments in all sections except $\S$~\ref{subsec:quantization}. Therefore, the latency and memory usage reported are models after 4-bit quantization. Further optimization in recent literature is thoroughly discussed in $\S$~\ref{subsection:review}.

\subsection{Overview}
\subsubsection{Inference Latency}
In  Figure~\ref{fig:model_size}, the inference latency including first token time and decode latency per token for the models ranging in size from 0.1B to 3B were measured, revealing that they can be categorized into three intervals: 0.1-1B, 1-2B, and 2-3B. The inference latency within each interval is relatively similar and aligns with the latency increase as the model size grows. 
For models of similar size from different architectures, the first token time during the prefill stage vary significantly. The first token time of Qwen2-0.5B is 1.46$\times$ of Qwen1.5-0.5B and is close to that of OpenELM-1.1B which has 2.18$\times$ model size. Qwen2 adopts an architecture that shares the embedding layer and the LM head, allowing more parameters to be allocated to the attention block, specifically to the attention mechanism and FFN, which are more computationally intensive and time-consuming. The latency of Pythia-1.4B is higher than that of SmolLM-1.7B, Qwen2-1.5B, Qwen1.5-1.8B, and Qwen-1.8B, despite these models being larger than Pythia-1.4B. A similar phenomenon is observed in larger models: among models in the 2-3B range, Phi-2 has 1.11$\times$ latency than the larger one OpenELM-3B.
To be noted, prefill stage is often regarded as the dominate phase in end-to-end LLM inference on devices, since on-device LLM tasks often involve long-context understanding for context-awareness or personalization need~\cite{xu2024empowering}.

However, the model's latency during the decode stage more closely follows a linear relationship with model size.
The latency of Qwen2-0.5B and Qwen1.5-0.5B get close. Unlike the prefill phase, Pythia-1.4B has a lower decode latency compared to larger models. Among the 2-3B models, Gemma-2B, Phi-2, and OpenELM-3B show a trend of latency positively correlating with model size.


\subsubsection{Memory Footprint}\label{subsubsec:memory footprint}
The evaluation of memory footprint in Figure~\ref{fig:model_size} used \texttt{llama.cpp} on Jetson. 
The size of models range from 0.1B to 3B parameters and the memory footprint range from 275MB to 2456MB. Due to llama.cpp defaulting to allocate KV cache and compute buffer according to the maximum context length of the model, models that support longer contexts end up consuming significantly more memory than others. In our experiments, we set the maximum context length for all models to 2048 to standardize memory usage.
Under the same context length, memory usage is linearly related to model size. However, some models exhibit memory usage that does not align with their size, such as Gemma-2B, Bloom-560M, and Bloom-1B1. These models have larger vocabularies compared to others: Gemma-2B has a vocabulary size of 256,000, while the Bloom series has a vocabulary size of 250,880. The OpenELM series has lower memory usage compared to models of similar parameter size for two reasons. First, it uses a vocabulary size of 32,000, smaller than the 50,000 used by most models. Second, it employs GQA, which reduces the KV cache, instead of MHA. We will explain in $\S$~\ref{subsubsec:memory breakdown} why vocabulary size has a significant impact on model memory usage.

\begin{remark}
\textbf{Insights}: We have three key insights from the overview of SLM runtime cost.
\begin{itemize}[leftmargin=*]
    \item Apart from the model size, the model architecture also impacts latency. Factors such as the number of layers, the width of the FFN, the size of the vocabulary, and whether parameters are shared play significant roles.
    For example, Qwen1.5-0.5B has 25.4\% more parameters than Qwen2-0.5B, but runs 31.9\% faster on Jetson Orin NX 16GB.
    The correlation is likely hardware-dependent.
    This indicates that SLM development shall be aligned with the hardware where it will be deployed.
    
    \item The impacts of model architecture on inference speed is more significant at prefill stage than decode stage. This is because that the computational density in the prefill stage is higher, making it more likely to be compute-bound, while the decode stage is primarily memory-bound. Differences in model architecture can more easily affect the compute-bound scenarios; for example, wider and shallower models have higher computational parallelism. 

    \item Runtime memory usage is generally linearly correlated with the model's parameter count. A few models have larger memory usage compared to others with similar parameter counts, typically due to their larger vocabulary sizes. For instance, the Bloom series has a vocabulary size of 250,880, which is 5$\times$ to 8$\times$ larger than that of most models.
\end{itemize}
\end{remark}

\subsection{Impact of Quantization and Hardware}
\subsubsection{Impact of Quantization}\label{subsec:quantization}
\input{figs-cost/fig_quan}
The benefits of quantization for reducing inference latency on server-side GPUs likely stem from three factors: higher computational throughput of Tensor Cores for int8 operations, reduced memory access overhead, and the decrease in heat generated by reduced memory access. On mobile devices, such as Jetson, support for int8 computation is lacking, but memory access overhead can still be effectively reduced. This reduction comes from data compression due to the lower precision of activation values and parameters, which in turn improves cache utilization.

We utilized five quantization methods to test the latency of Phi-1.5, as shown in Figure~\ref{fig:model_quan}.
Qn\_K (and Qn\_K\_M) refer to the quantization of a model to \textit{n} bits using the \textit{k}-quants method with a medium (M) number of parameters, while Qn\_0 specifically refers to symmetric quantization of a model to \textit{n} bits. For the prefill phase, when the prompt length is relatively short, quantization can reduce latency by at least 25\%. However, this benefit diminishes as the prompt length increases. When the prompt length approaches 50, the Q6\_K and Q3\_K quantization methods result in latency that is nearly identical to, or even exceeds, that of the unquantized FP16 model. On the other hand, the Q8\_0, Q4\_K\_M, and Q5\_K methods provide stable performance improvements. Among these, Q4\_K\_M performs the best, reducing latency by an average of 50\%. Quantization during the decode stage delivers more consistent performance gains, reducing decode latency by up to 75\% and no less than 17\%. As in the prefill stage, the Q4\_K\_M method proves to be the most effective, while Q6\_K remains the least efficient.
\begin{remark}
\textbf{Insights}: We have two key insights about the impact of quantization methods for latency with different prompt length and output token length.
\begin{itemize}[leftmargin=*]
    \item The benefits of quantization during the decode stage are greater than those in the prefill stage. On mobile devices, quantization mainly reduces memory access overhead. Since the decode stage is more bandwidth-bound, it gains more from quantization compared to the compute-bound prefill stage. 

    \item The benefits of quantization during the prefill stage decrease with prompt length increasing. Quantization compresses weights and kv cache. All tokens share the weight file so the benefit for each token decreases when there are more tokens in prompt. However, decode stage has more memory operation in kv cache so it still is benefited with generation tokens increasing.
    
    \item More regular quantization precision leads to better performance. Although 3-bit quantization offers a higher model compression rate, 4-bit quantization performs better in both the prefill and decode stages. The inferior performance of 3-bit quantization is due to its irregular bit-width, which lacks hardware optimization support and incurs additional overhead from data alignment and padding. As a result, despite its lower compression rate, 4-bit quantization is more efficient. Similarly, irregular 5-bit and 6-bit quantization result in inference latency that is comparable to, or even higher than 8-bit quantization, despite offering higher compression rates.

\end{itemize}
\end{remark}

\subsubsection{Impact of Hardware}
\input{figs-cost/fig_hardware}
\input{figs-cost/fig_mobile}
We conducted tests using Bloom-1B1 on two types of edge devices: the Jetson Orin NX 16GB, which utilizes its GPU, and the Meizu 18 Pro, which relies on its CPU. During the prefill phase, for a single token, the Jetson is approximately 10 to 20 times faster than the Meizu 18 Pro. Both the Jetson and the Meizu 18 Pro show a linear increase in first token time as the prompt length increases, with the Jetson's advantage becoming more obvious as the prompt length grows. During the decode phase, the latency per token increases as the number of generated tokens grows. On the Meizu 18 Pro, the latency rises sharply from 1 to 10 tokens and then levels off after 10 tokens. This initial steep rise in latency from 1 to 10 tokens is due to the temperature increase, which triggers the Dynamic voltage and frequency scaling (DVFS) or thermal throttling to adjust power consumption and frequency, thereby reducing computational efficiency. In contrast, the Jetson, benefiting from a more effective cooling system, shows noticeable fluctuations and increases in latency only after 30 tokens.

We also tested the prefill and decode times of Qwen1.5-1.8B on three smartphones. To minimize the impact of power consumption, a 60-second interval was set between each test. As shown in the figures, the latency for prefill and decode on the three smartphones increases linearly with the number of tokens. The Xiaomi 12S performed the best with the lowest latency, highlighting the efficiency of the Snapdragon 8Gen1+ chip. The Pixel 7 Pro followed, delivering competitive performance, while the MEIZU 18 Pro had the highest latency due to its older Snapdragon 888 chip and lower memory configuration.
\begin{remark}
\textbf{Insights}: We have two key insights about the impact of hardware for latency with different prompt length and output token length.
\begin{itemize}[leftmargin=*]
    \item 
    GPU shows an even greater advantage over the CPU during the prefill phase. The prefill phase involves parallel processing of tokens within the prompt, whereas the decode phase generates each token sequentially. Therefore, the prefill phase has a higher degree of parallelism, making it more suitable for GPUs, which have more parallel computing units.
    \item 
    The Jetson demonstrates better performance stability compared to the smartphone. Due to its relatively simple hardware structure, which facilitates better heat dissipation, the Jetson maintains more stable latency during lengthy inference tasks.
    \item 
    The development of System on a Chip (SoC) generations effectively improves inference efficiency. 
\end{itemize}
\end{remark}

\subsection{Latency and Memory Breakdown}\label{subsection:breakdown}
\subsubsection{Latency Breakdown}\label{subsubsec:latency breakdown}
\input{figs-cost/fig_latency_breakdown}
In Figure~\ref{fig:breakdown_latency}, we conducted a breakdown analysis of Qwen2-0.5B and Qwen1.5-0.5B, models with the similar size but different latency, and measured the time distribution across the Embedding, Attention, FFN(Feed-Forward Network), and LM\_Head. For Qwen1.5 and Qwen2, the prefill phase is predominantly characterized by the high involvement of the Attention and FFN layers. In Qwen1.5, the Attention layer has a slightly higher proportion than the FFN layer, whereas in Qwen2, the FFN layer's contribution is noticeably greater than that of the Attention layer. This is due to Qwen2 having a wider FFN layer compared to Qwen1.5. During the decode phase, the proportion of the Attention layer in Qwen1.5 increases, which could be attributed to the increased length of the KV (Key-Value) Cache. As for Qwen2, it still has longest time for FFN.

We also conducted an operator breakdown analysis on Qwen1.5-0.5B and Qwen2-0.5B. Regardless of the model or whether it is during the prefill or decode phase, the operator mul\_mat\_vec\_q, which represents matrix-vector multiplication, accounts for over 80\% of the time. The mul\_mat\_vec\_q operator in Qwen2-0.5B has a higher proportion compared to Qwen1.5-0.5B, which may also be due to its wider FFN layer.
\subsubsection{Memory Breakdown}\label{subsubsec:memory breakdown}
\input{figs-cost/fig_memory_breakdown}

In $\S$~\ref{subsubsec:memory footprint}, we found that in addition to model size, vocabulary size also has a significant impact on memory usage. In the figure, we provide a breakdown analysis of the model's memory usage. The runtime memory is primarily consumed by model parameters, the KV cache, and intermediate variables during computation.
Models with larger vocabularies require a larger compute buffer due to the need for a matrix of hidden\_size * vocabulary\_size in the Output Layer. Bloom series have a 250880 vocabulary. As Figure~\ref{fig:breakdown_of_memory_voca}, Bloom-560M's compute buffer size is 492MB, which is 3.5$\times$ larger than that of the larger OpenELM-1.1B with vocabulary size of 32000. Similarly, Bloom-1B1's compute buffer size is 496MB, which is 1.6$\times$ larger than that of the larger Qwen2-1.5B with vocabulary size of 151936. Models using GQA tend to have smaller KV Caches compared to those using Multi-Head Attention (MHA). For instance, OpenELM-3B's KV Cache size is 164MB, which is 3.9$\times$ smaller than that of StableLM-zephyr-3B.

When the input context length is long, the sizes of the Compute Buffer and KV Cache become the primary determinants of the model's memory usage as we seen in Figure~\ref{fig:breakdown_of_memory_context}. For the Qwen2 series of models, when the context length reaches its upper limit 131072, the Compute Buffer and KV Cache occupy between 83\% and 87\% of the total memory. For Qwen1.5 with max context length 32768, the Compute Buffer and KV Cache occupy between 85\% and 90\% of the total memory.
\begin{remark}
\textbf{Insights}: We have two key insights for the breakdown of inference latency and memory footprint.
\begin{itemize}[leftmargin=*]
    \item \texttt{Mul\_mat\_vec} (matrix by vector multiplication) is the most time-consuming operations of SLM, which constitute more than 70\% end-to-end inference time. 
    
    \item Context length is crucial for model runtime memory usage. When context length gets to 32,000, the KV cache will take up over 80\% memory.
     
\end{itemize}
\end{remark}



 \subsection{Optimizations for On-device Deployment}\label{subsection:review}
 Since SLMs are deployed on resource-constrained devices, numerous approaches have been developed to optimize their performance in terms of latency, memory, and other overheads. These methods can be categorized into two types: Online and Offline. Online optimization focuses on enhancing SLM performance during runtime, with key techniques including hardware-aware optimizations, model collaboration and quantization. In contrast, Offline optimization targets SLM improvements prior to deployment, including optimizing datasets, model architecture and knowledge distillation.
 \subsubsection{Online Optimization}
Hardware-aware optimization requires fully leveraging the heterogeneous computing capabilities of edge devices. This involves optimizing task scheduling across CPUs, GPUs, and NPUs, as well as effectively utilizing multiple storage hierarchies, such as cache, DRAM, and flash.
Earlier frameworks were primarily designed for CPU-GPU orchestration.
EdgeNN~\cite{zhang2023edgenn} optimizes inference on IoT devices with a CPU-GPU architecture by improving the use of unified memory and enhancing task scheduling between the CPU and GPU, bringing an average of 3.97×, 3.12×, and 8.80× speedups to inference on the CPU of the integrated device, inference on a mobile phone CPU, and inference on an edge CPU device, respectively.
Transformer-Lite~\cite{li2024transformer} is a framework for efficiently deploying large language models on mobile phone GPUs by optimizing memory usage and computational efficiency. It utilizes techniques like model pruning, quantization, and layer fusion to reduce memory footprint and computational load, achieving up to 5.7x faster inference compared to CPU-based methods while maintaining model accuracy.
PowerInfer~\cite{song2024powerinfer} introduces an efficient framework for serving large language models (LLMs) on consumer-grade GPUs. It optimizes inference by partitioning the model into cold-activated neurons and hot-activated neurons, and assigns the heavy computation tasks to the GPU while offloading lighter tasks to the CPU. This results in up to 2.4x faster performance, reduced memory usage, and lower power consumption, making LLM inference feasible on consumer-grade hardware.

Recent systems like PowerInfer-2~\cite{xue2024powerinfer} and mllm-NPU~\cite{xu2024empowering} are specifically designed for off-the-shelf modern smartphones, where the AI computing power resides in the NPU. Consequently, both frameworks focus on optimizing the CPU-NPU architecture.
PowerInfer-2, as an extension of the original PowerInfer, adopts a neuron cluster approach, where matrix operations are decomposed into clusters of neurons. Based on the activation sparsity, it schedules the computation on either the NPU or CPU. This efficient allocation of resources results in up to a 27.8x speed improvement compared to existing state-of-the-art frameworks. Notably, PowerInfer-2 can serve a 47B parameter LLM at an impressive rate of 11.68 tokens per second, enabling fast and efficient LLM inference on smartphones without significant accuracy degradation.
mllm-NPU optimizes at three levels—prompt, tensor, and block and achieves up to 22.4x faster prefill speed, 30.7x energy savings, and up to 32.8x speedup in end-to-end real-world applications, with the ability to process over 1,000 tokens per second for Qwen1.5-1.8B.
RIPPLE~\cite{wang2024ripple} proposes the
concept of Neuron Co-Activation, where neurons frequently activated together are linked to facilitate continuous read access and optimize data transfer efficiency.
It achieves up to 5.93× improvements in I/O latency compared to the state-of-the-art.

For the memory optimization, LLMS~\cite{yin2024llm} has proposed an effective solution. To reduce the context switching overhead when running LLMs under tight device memory constraints, LLMS employs fine-grained memory management by decoupling the memory management of the application and LLM. It utilizes globally optimized key-value (KV) cache compression and swapping techniques to efficiently manage memory resources. LLMS achieves up to 100x faster performance in context-switching compared to existing solutions on various edge devices.

The deployment of mixture-of-experts (MoE) models on edge devices also faces challenges in terms of huge memory usage.
EdgeMoE~\cite{yi2023edgemoe} is the first on-device inference engine for MoE-based large language models, optimizing memory and computation by storing non-expert weights in device memory and expert weights in external storage, fetched only when activated. It further improves efficiency through expert-wise bitwidth adaptation and advanced expert management techniques. It reduces memory footprint by 1.05x to 1.18x compared to traditional models that store the entire model in memory. Additionally, EdgeMoE demonstrates 1.5x to 2.5x faster inference on Jetson TX2 (GPU) and Raspberry Pi 4B (CPU), with some tasks showing up to 10x speedup due to optimized memory and computation management.
MoE Cache-Conditional~\cite{skliar2024mixture} also chooses to keep non-expert weights resident in DRAM, while expert weights are dynamically scheduled from flash storage back to DRAM.
It present on-device results demonstrating 2× speedups on mobile devices.

Speculative sampling is commonly used on the cloud to reduce the computational overhead of LLMs. This idea of model collaboration can similarly be applied effectively on SLMs.
LLMCad~\cite{xu2023llmcad} is an efficient on-device inference engine for generative NLP tasks on mobile devices. By combining a smaller memory-resident LLM for token generation and a larger LLM for token validation, LLMCad achieves up to 9.3x faster performance compared to existing engines on IoT devices and smartphones, respectively, without comprising accuracy.

Besides validating the outputs of smaller models with large models, models of different scales can also be applied to different tasks. ELMS~\cite{yin2024elms} proposes a one-time neuron reordering technique for efficient sub-model generation and a dual-head compact language model for elastic mobile LLM services. ELMS improves accuracy by up to 16.83\% on end-to-end traces and speeds up inference time by up to 5× with less than 1\% switching overhead, while maintaining comparable memory usage and utilizing fewer than 100 GPU hours for model preparation.

 \subsubsection{Offline Optimization}
Datasets play a crucial role in the accuracy of SLMs. Recent studies have focused on enhancing model performance by optimizing data utilization or generating specific types of data for training. TinyStories~\cite{eldan2023tinystories} is a synthetic dataset designed for training small language models. Trained on those child-friendly vocabulary, models as small as 28M parameters can produce coherent stories comparable to GPT-2-XL (1.5B parameters). Models trained on TinyStories require less than one GPU-day for training while exhibiting emergent reasoning and language capabilities.
AS-ES learning~\cite{xi2024learning} is a novel data-efficient training paradigm for improving the reasoning performance of small models by segmenting chain-of-thought data into Abstractive Segments and Extractive Segments. 
This method enables iterative generation without requiring additional data or modifications to the model architecture. On math word problems, AS-ES learning improves accuracy by up to 15.28\%, while reducing BLEU scores, suggesting a trade-off that favors logical reasoning over memorization. For PET scan summarization tasks, it achieves up to 30.6\% higher ROUGE-L scores, highlighting its effectiveness across diverse reasoning-intensive tasks.
Self-AMPLIFY~\cite{bhan2024self} improves small language models' reasoning ability by generating rationales using post hoc explanation methods, eliminating the need for auxiliary models or human annotations

Model architecture optimization involves enhancing the model's efficiency and performance by adjusting or redesigning its architecture.
OnceNAS~\cite{zhang2024oncenas} is a neural architecture search method designed for on-device inference on edge devices with constrained resources. By leveraging a continuous latent space for architecture representation and integrating parameter count, latency, and accuracy as optimization objectives, OnceNAS achieves a 10.49× size reduction and a 5.45× speedup compared to baseline methods. Evaluations across various edge devices, including Raspberry Pi and FPGA, demonstrate superior efficiency and generalizability, making OnceNAS highly suitable for edge intelligence applications like autonomous driving and smart healthcare.
Weight-Inherited Distillation (WID)~\cite{wu2023weight} proposes a task-agnostic BERT compression technique that eliminates alignment losses by directly inheriting weights through structural re-parameterization with row and column compactors. WID achieves up to 49.2\% parameter reduction while retaining 98.9\% of BERT-base performance on GLUE benchmarks and significantly outperforms baselines like TinyBERT and MiniLM in both accuracy and efficiency, with training time reduced by over 50\%.

Knowledge Distillation transfers knowledge from larger models to smaller ones through knowledge distillation, thereby improving the performance of smaller models, particularly in reasoning tasks. DISTILLM~\cite{ko2024distillm} proposes an efficient knowledge distillation framework, leveraging a novel skew Kullback-Leibler divergence loss and an adaptive off-policy approach to address training-inference mismatches and enhance computational efficiency. Compared to existing KD methods, DISTILLM achieves up to 4.3× faster training speeds while maintaining or improving performance on OpenLLaMA2-3B.
~\cite{magister2022teaching} explores transferring reasoning capabilities from large language models to smaller ones via chain-of-thought knowledge distillation. By fine-tuning smaller models like T5 on CoT outputs generated by LLMs such as PaLM 540B and GPT-3 175B, the authors achieve substantial improvements in reasoning tasks. For instance, T5 XXL’s accuracy on the GSM8K dataset increased from 8.11\% to 21.99\% when fine-tuned on CoT data generated by PaLM 540B, highlighting a 170\% improvement. The approach also demonstrates robustness across arithmetic, commonsense, and symbolic reasoning tasks.

Quantization is another effective method for model compression and reducing operational overhead, but there are few methods specifically designed for edge devices. Activation-aware Weight Quantization (AWQ)~\cite{lin2024awq} enables low-bit weight-only quantization for LLMs on edge devices. AWQ protects 1\% of the most critical weights using per-channel scaling informed by activation patterns, achieving a 3-4× speedup over FP16 implementations and reducing memory usage by 4×. AWQ is paired with TinyChat, a framework that further optimizes on-device inference, achieving up to 3.3× acceleration on GPUs.
~\cite{van2023fp8} compares FP8 and INT8 formats for efficient on-device deep learning inference. It demonstrates that while FP8 may offer slightly improved accuracy for certain outlier-dominated tasks, INT8 achieves superior overall efficiency, reducing hardware costs by up to 50\% and energy usage by 53\%. For post-training quantization and quantization-aware training, INT8 consistently outperforms FP8 in networks with well-behaved distributions, proving more robust and computationally efficient across a range of tasks.


%% file: figs-cost/fig_model_size.tex

\begin{figure}[t]
\centering
\includegraphics[width=0.99\linewidth]{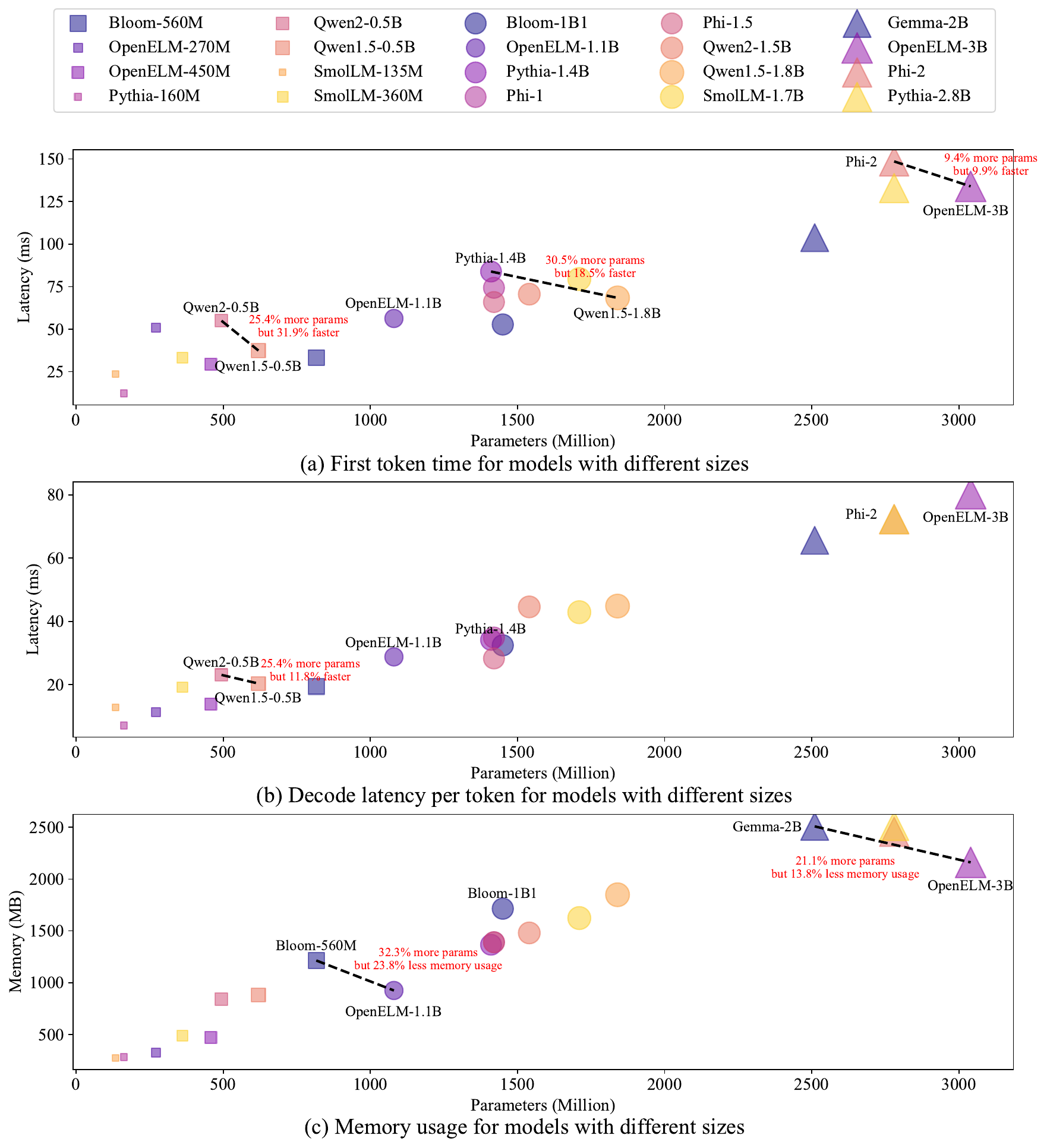}
    \caption{
        Latency and memory overview.
    }
    \label{fig:model_size}
\end{figure}

%% file: figs-cost/fig_quan.tex
\begin{figure}[t]
    \centering
    \begin{minipage}[b]{0.48\textwidth}
        \includegraphics[width=\textwidth]{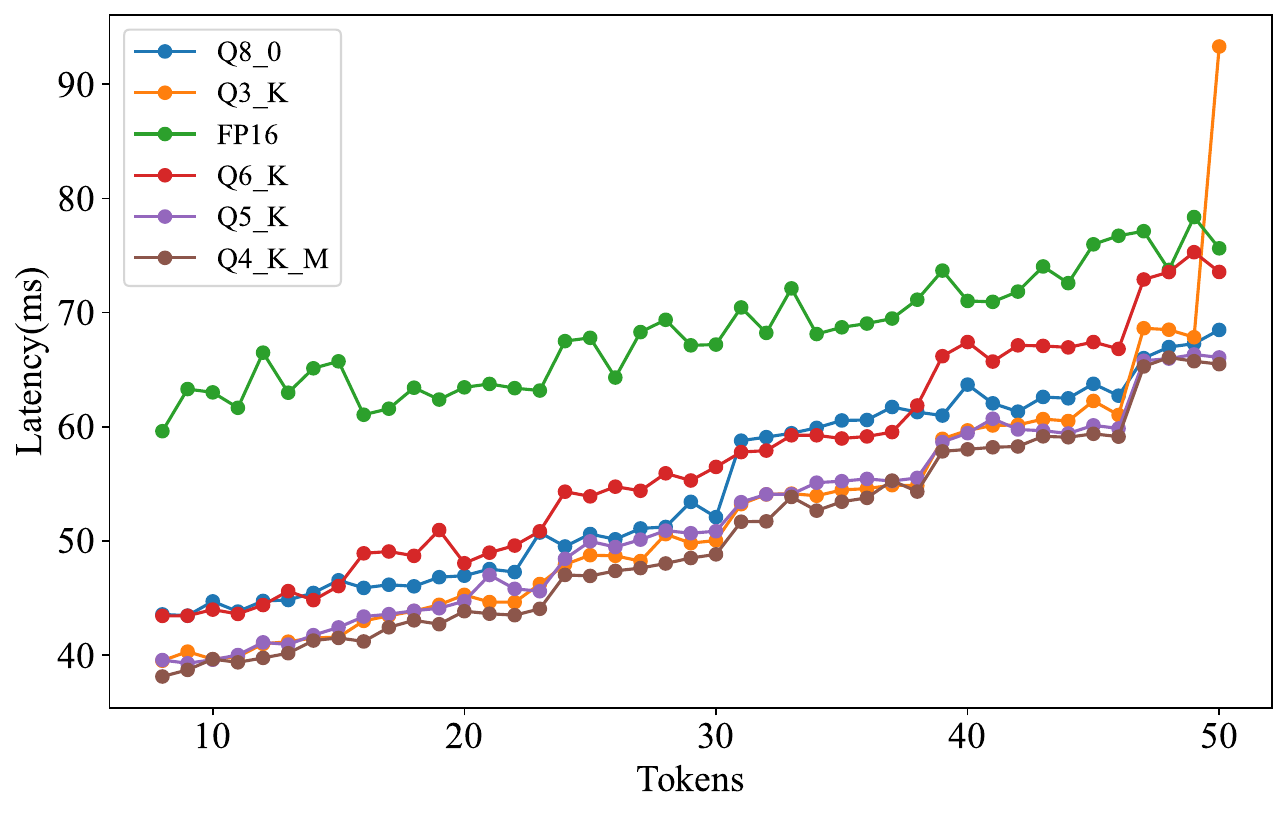}
        \subcaption{First Token Time}
    \end{minipage}
    \hspace{0.0\textwidth} 
    \begin{minipage}[b]{0.48\textwidth}
        \vspace{-4pt} 
        \includegraphics[width=\textwidth]{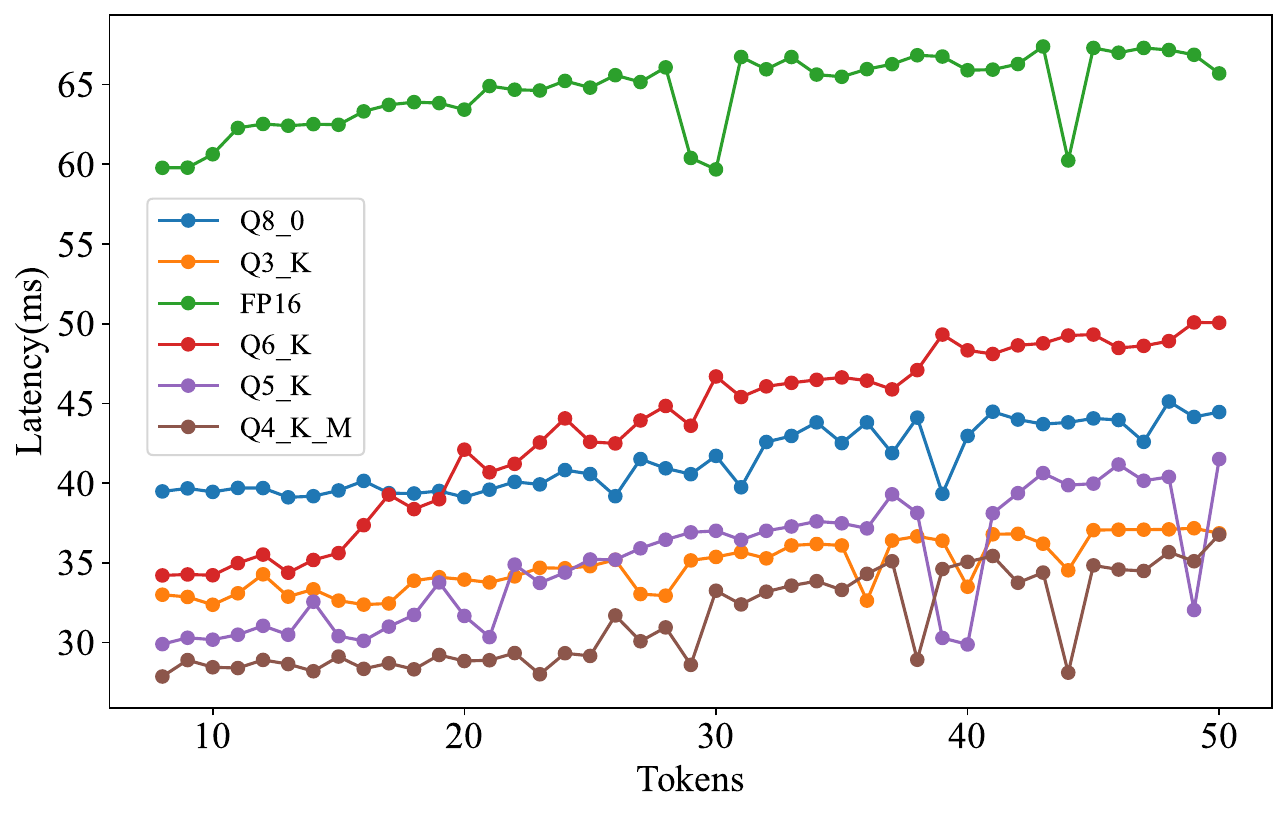}
        \subcaption{Decode Latency per Token}
    \end{minipage}
    \vspace{-4pt}	
    \caption{
        The relationship between the latency and quantization methods
    }
    \label{fig:model_quan}
\end{figure}

%% file: figs-cost/fig_hardware.tex
\begin{figure}[t]
    \centering
    \begin{subfigure}[b]{0.48\textwidth}
        \includegraphics[width=\textwidth]{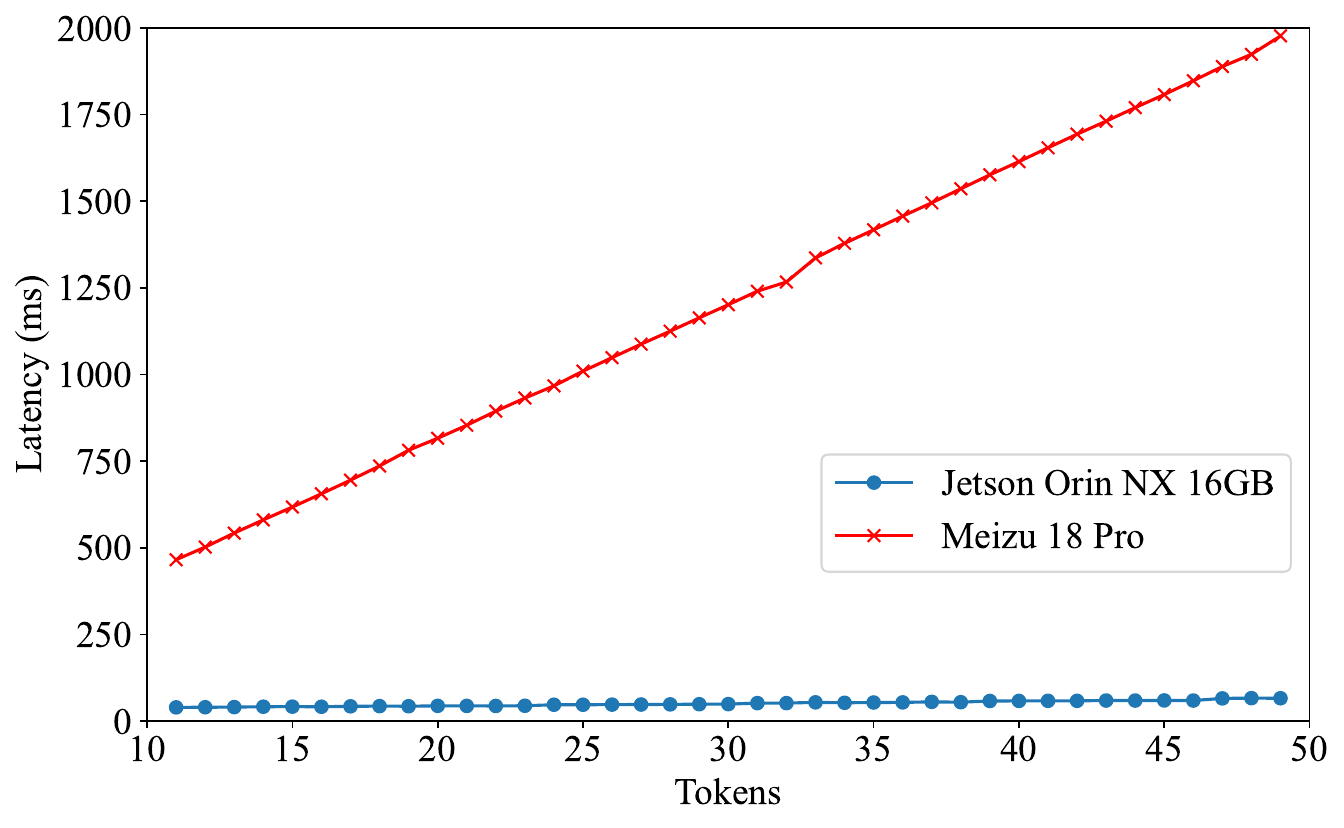}
        \caption{Prefill}
        \label{fig:prefill_time}
    \end{subfigure}
    \begin{subfigure}[b]{0.48\textwidth}
        \includegraphics[width=\textwidth]{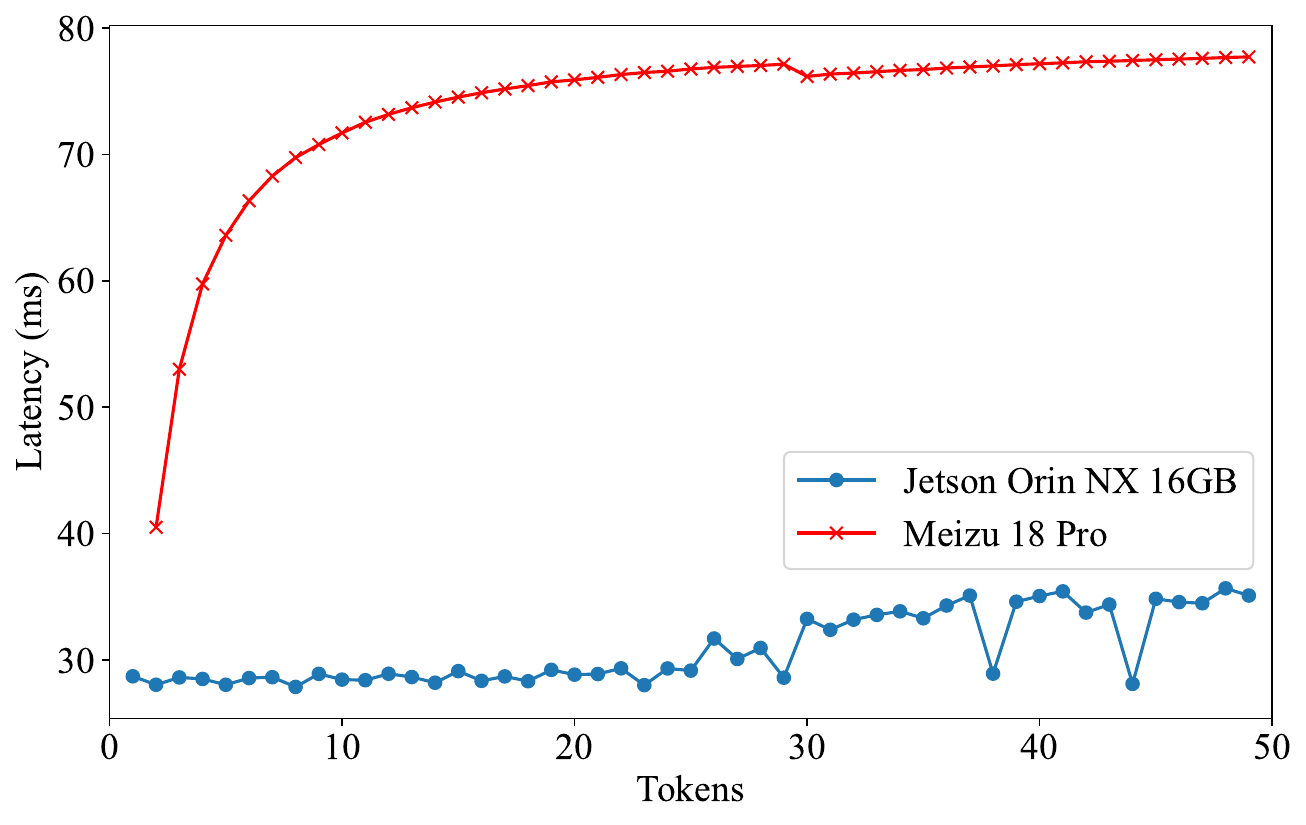}
        \caption{Decode}
        \label{fig:decode_time}
    \end{subfigure}
    \caption{Latency on GPU and CPU.}
    \label{fig:latency}
\end{figure}

%% file: figs-cost/fig_mobile.tex
\begin{figure}[t]
    \centering
    \begin{subfigure}[b]{0.48\textwidth}
        \includegraphics[width=\textwidth]{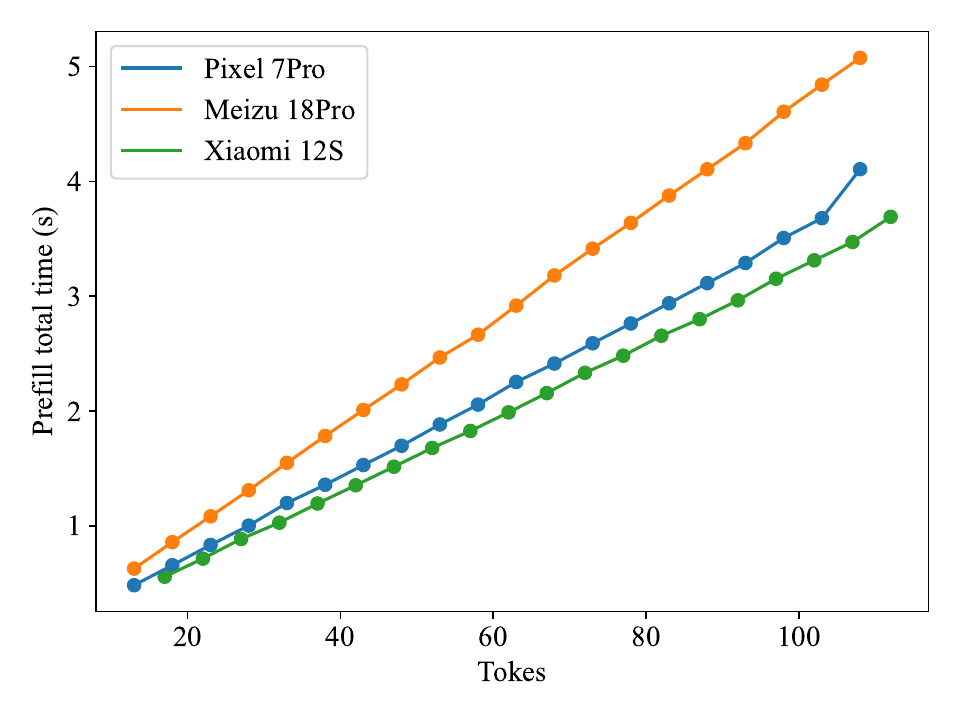}
        \caption{Prefill}
        \label{fig:prefill_time}
    \end{subfigure}
    \begin{subfigure}[b]{0.48\textwidth}
        \includegraphics[width=\textwidth]{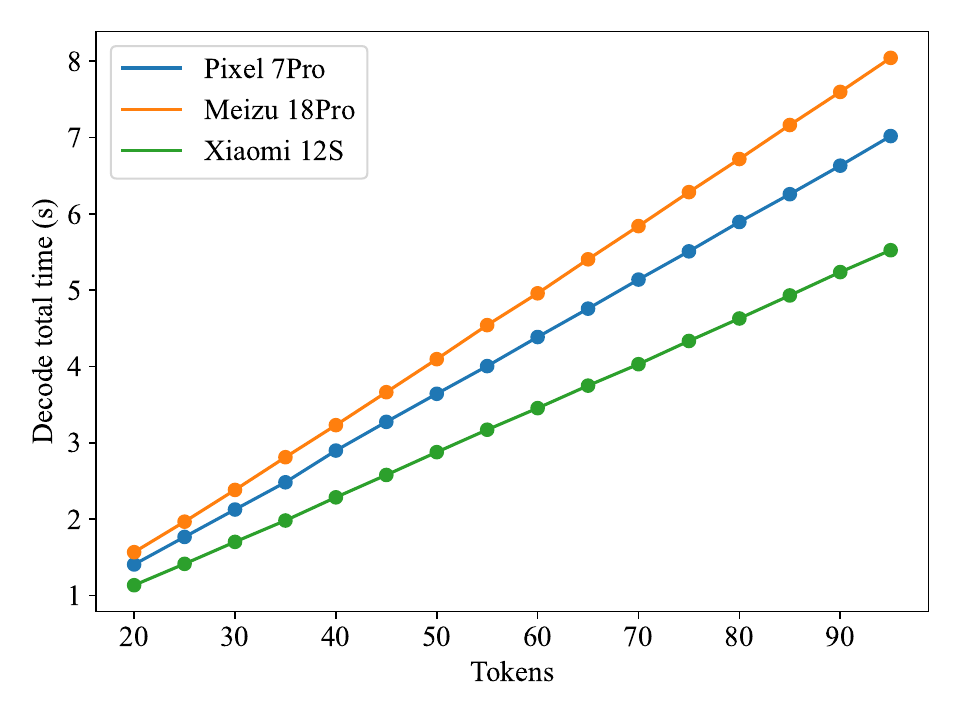}
        \caption{Decode}
        \label{fig:decode_time}
    \end{subfigure}
    \caption{Latency on different smartphones.}
    \label{fig:latency}
\end{figure}

%% file: figs-cost/fig_latency_breakdown.tex
\begin{figure*}[t]
    \centering
    \begin{subfigure}[b]{0.45\textwidth}
        \includegraphics[width=\textwidth ]{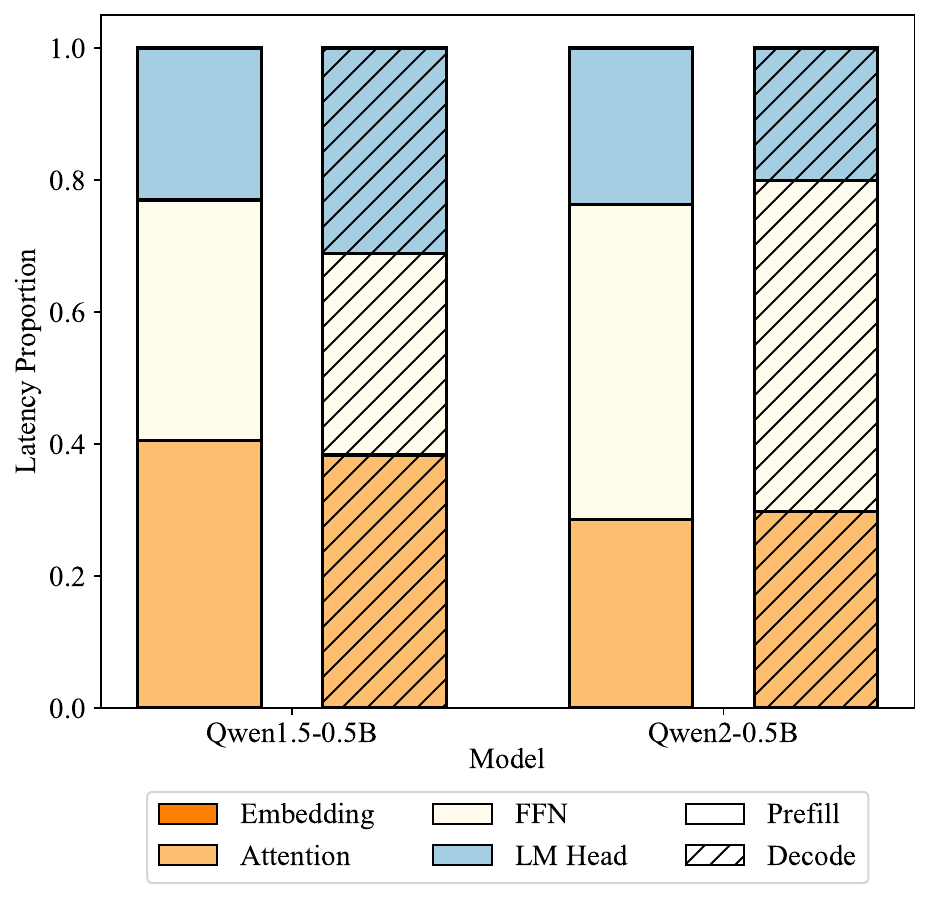}
        \caption{Layer granularity}
        \label{fig:breakdown_of_time_by_layer}
    \end{subfigure}
    \hfill
    \begin{subfigure}[b]{0.45\textwidth}
        \includegraphics[width=\textwidth]
        {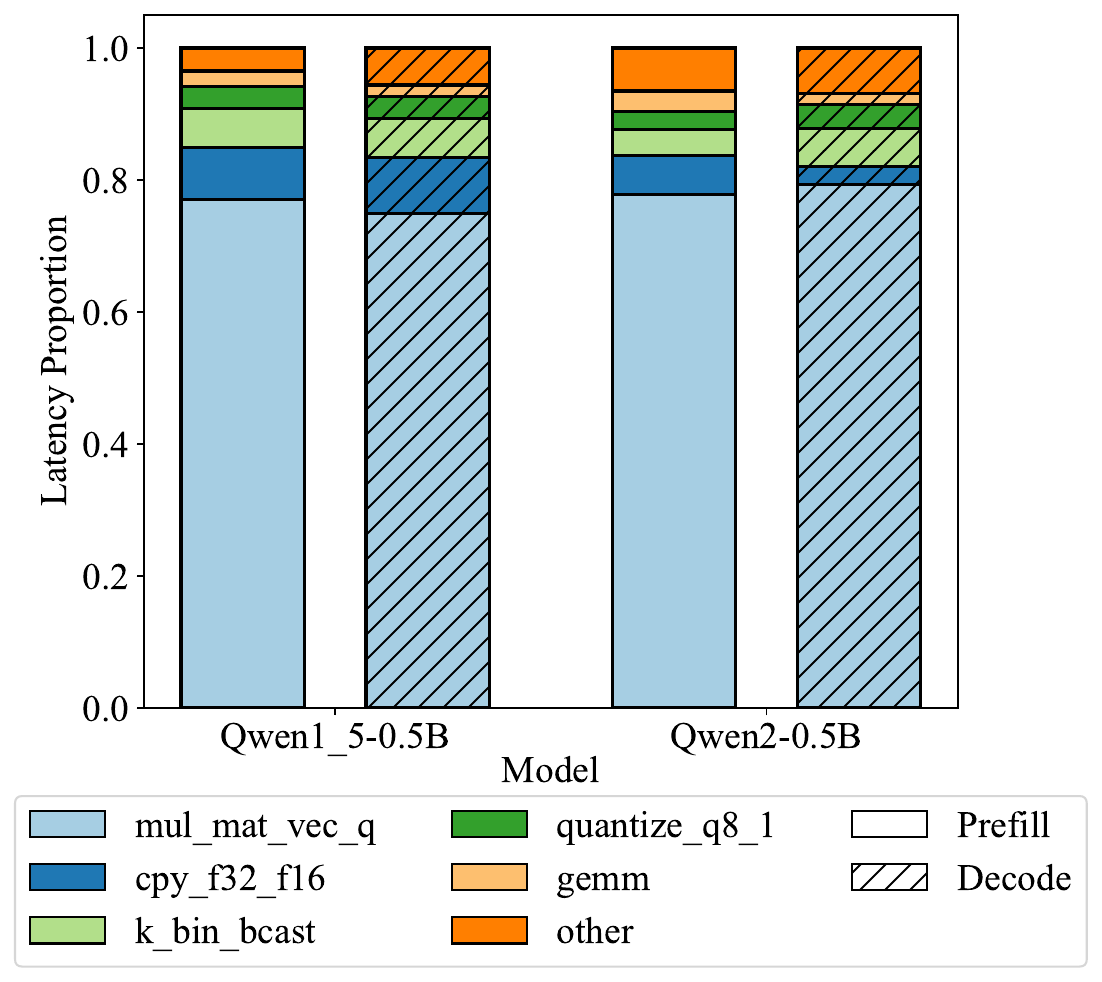}
        \caption{Op granularity}
        \label{fig:breakdown_of_op}
    \end{subfigure}
    \caption{On-device inference latency Breakdown.}
    \label{fig:breakdown_latency}
\end{figure*}

%% file: figs-cost/fig_memory_breakdown.tex
\begin{figure*}[t]
    \centering
    \begin{subfigure}[b]{0.49\textwidth}
        \includegraphics[width=\textwidth ]{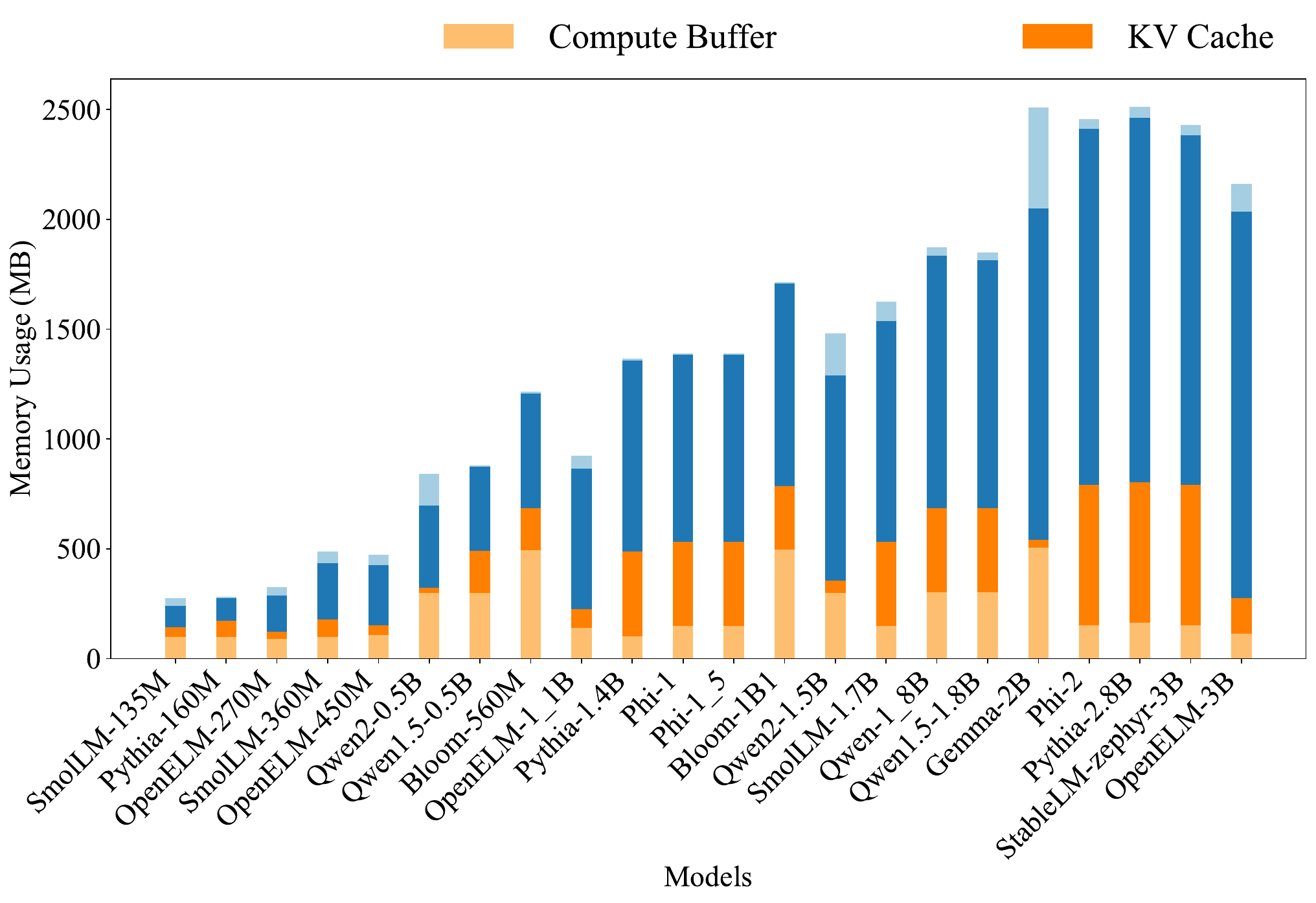}
        \caption{Context length is 2048.}
        \label{fig:breakdown_of_memory_voca}
    \end{subfigure}
    \hfill
    \begin{subfigure}[b]{0.49\textwidth}
        \includegraphics[width=\textwidth]
        {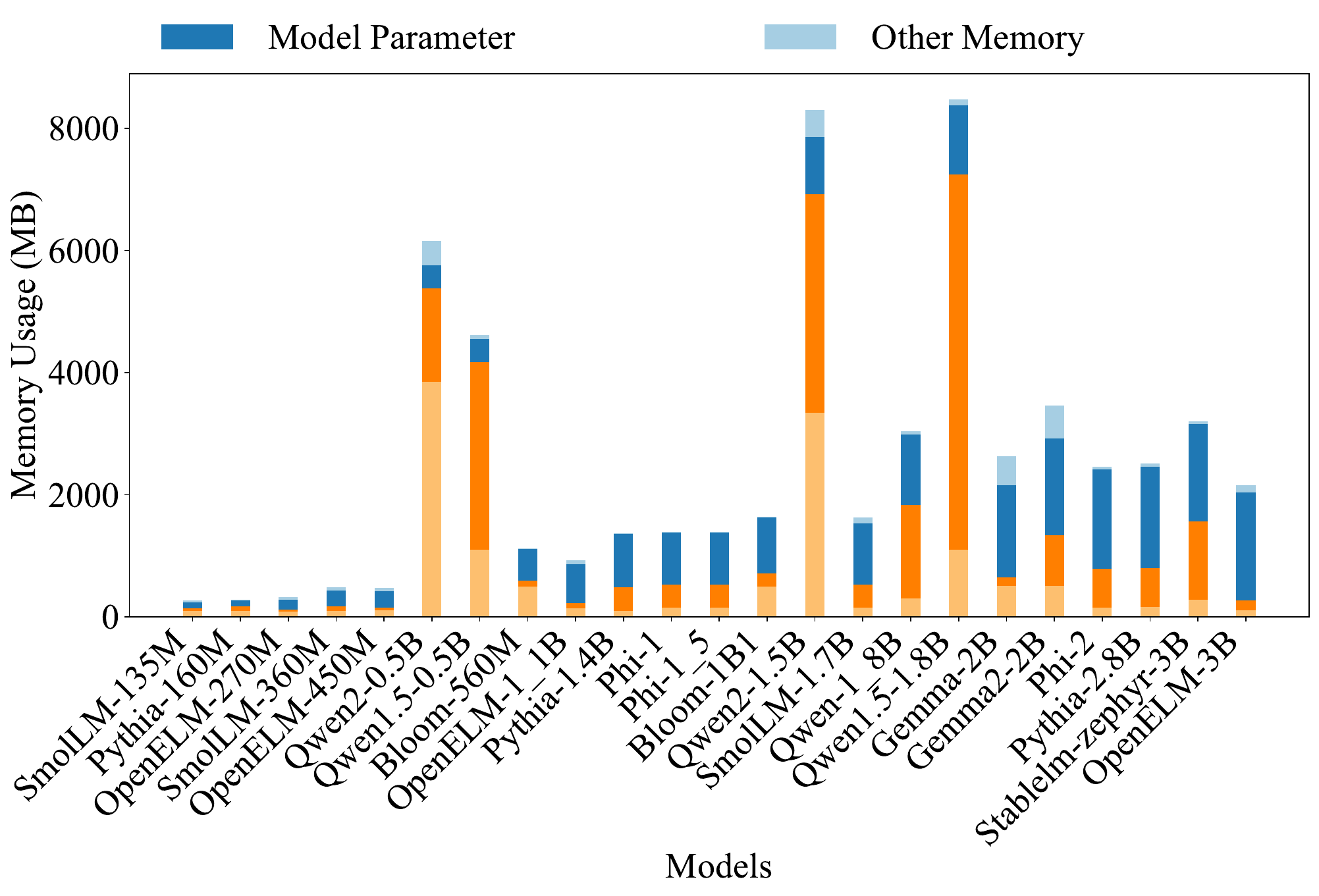}
        \caption{Context length is max context length of models.}
        \label{fig:breakdown_of_memory_context}
    \end{subfigure}
    \caption{Memory Breakdown.}
    \label{fig:breakdown_memory}
\end{figure*}

%% file: sec-conclusion.tex
\section{Conclusions and Future Directions}

This paper makes a comprehensive survey and measurement to small language models (100M--5B parameters), including their capabilities, runtime cost on devices, and innovations.
We then summarize the key insights to inspire future research on small language models. Specifically, we expect following directions worthy explorations.

\textbf{Co-design and co-optimizations of SLM architecture and device processors.}
With given parameter size, the concrete SLM architecture still has huge impacts on the runtime speed, as discussed in $\S$\ref{sec:cost}.
This includes both the basic transformer configuration (e.g., depth-width ratio, attention type, activation) and how efficiently they can be quantized for execution on integer-optimized processors like NPUs~\cite{xu2024empowering}.
To push the limit of SLMs towards optimal accuracy-speed tradeoff, we advocate for extreme co-design and optimizations of SLM architecture with specific device hardware, possibly searching for a speed-optimal architectures before pre-training on them.

\textbf{Constructing high-quality synthetic dataset.}
Two recent pre-training datasets, DCLM and FineWeb-Edu, have exhibited superior performance and greatly closed the gap between SLMs trained on open/closed datasets.
The key innovation of them is to use carefully trained model to filter out high-quality data portion from a large corpora.
We believe that we are still at the very beginning of such synthetic data research, and the space to be explored remains huge.
It is urgent to standardize a process of synthetic data curation (deduplication, filtering, mixing, evaluation, etc).

\textbf{A deployment-aware Chinchilla law for model scaling.}
As discussed in $\S$\ref{subsec:dataset}, there is a notable trend to ``over-train'' SLMs on large amount of tokens as compared to what is instructed by Chinchilla law.
This is because SLMs are to be deployed on resource-constrained devices, where the device memory and compute are the primary constraining factors, instead of the training-time FLOPs.
Such strategy turns out to be effective to certain extent.
However, the training data size cannot be scaled out infinitely, and it still remains an open question on how to determine the optimal data scaling method for SLMs.
Intuitively, the decision relies on not only the training and inference cost, but also the lifecycle of SLM deployment and the economic benefits it is estimated to bring out with more training data.
If sparsity is taken into consideration (e.g., MoE), the question is further complicated.

\textbf{Continual on-device learning for personalization.}
Deployed on devices, SLMs are able to access on-device data to achieve better performance or personalization, without concerns on data leakage.
There are two approaches in general.
One is injecting personal data into prompts, using retrieval-augmeted generation technique.
This approach, however, can significantly increase the on-device cost (text embedding generation, longer prompt processing), and require the data to be stored on devices longer for retrospective query.
The second approach simply uses the data to finetune the SLM, so the knowledge is embedded into the weights and the data can be discarded after finetuning.
Though, such an approach faces critical challenges as well, especially the huge resource demand (memory and energy footprint) of on-device SLM training even with parameter-efficient finetuning techniques.
One possible direction is to apply zeroth-order optimization to SLM finetuning~\cite{xu2023fwdllm}, to avoid storing the activations in memory and be compatible with inference-time hardware accelerators (e.g., mobile NPU).

\textbf{Device-cloud SLM-LLM collaboration.}
While our measurements demonstrate that SLM capability is fast evolving, the gap between it and the cloud-side LLM will exist.
To achieve full-scale intelligence while not comprising privacy and availability (much), device-cloud collaboration will become an important research topic~\cite{chen2024role,zhang2024fast}.
Intuitively, the SLM can be used as a ``filter'' that solves the easy tasks confidently on devices, and cloud LLM can be treated as a safe guard for critical, difficult tasks.
This approach bottlenecks at the capability of SLM and the decision module of what tasks are easy enough for SLMs.
Does a better collaboration approach exist?
Such research is challenged by the auto-regressive manner of casual language models.

\textbf{Benchmarking SLMs fairly.}
First, SLMs are known to have systematic overfitting issue~\cite{zhang2024careful} to widely-used benchmarks such as GSM8k.
Given most SLMs (especially those state-of-the-art ones) are trained on closed dataset, it becomes challenging to fairly compare their capability.
Second, SLMs are designed to be deployed on devices, where the target tasks could differ from those hosted in clouds.
There have been limited efforts in constructing a comprehensive capability and performance benchmark for SLMs.
For example, when deployed on smartphones, SLMs are more likely to handle tasks that are sensitive to user data, e.g., auto-reply based on historical chatting text, or GUI context understandings.
Such ``ad-hoc'' tasks are not included in the existing LLM benchmarks commonly used, thereby their importance is often underrepresented.

\textbf{Sparse SLMs.}
During investigation SLMs, we find very little study of sparse SLMs, neither at architecture level (e.g., mixture-of-experts or MoE) or runtime level (e.g., activation sparsity).
The reasons could be multifold.
First, SLMs are supposed to have lower sparsity level as compared to LLMs~\cite{song2024achieving}, thereby the benefits in exploiting the sparsity for speedup or memory saving could be limited.
Second, architecture-level pre-assumed sparsity method like MoE is often considered to sacrifice memory usage for less computing intensity~\cite{artetxe2021efficient,krajewski2024scaling}, which does not fit memory-constrained devices.
One way to break down the memory wall for sparse SLMs is to leverage the external storage on devices, e.g., flash on smartphones.
By placing ``cold weights'' on storage and retrieve them on demand, SLMs can be scaled out to larger capacity~\cite{alizadeh2023llm,yi2023edgemoe,xue2024powerinfer}.
The challenges are to hide the I/O latency through careful pipeline, and keep compatible with heterogeneous hardware accelerators.